\pdfoutput=1
\PassOptionsToPackage{hyperfootnotes=false}{hyperref}
\RequirePackage{fix-cm}
\documentclass[11pt,a4paper,onecolumn]{personalarxiv}
\usepackage{mathtools,array,tikz}
\usetikzlibrary{arrows.meta,positioning,fit,calc}

\makeatletter
\let\paper@original@xfloat\@xfloat
\def\@xfloat#1[#2]{\paper@original@xfloat{#1}[#2p]}
\makeatother

\newcommand{\V}{\mathcal V}
\newcommand{\E}{\mathbb E}

\newcommand{\base}{\textsc{baseline}}
\newcommand{\ours}{\textsc{LangDiT }}

\DeclareMathOperator{\RMSNorm}{RMSNorm}
\newcolumntype{Y}{>{\raggedright\arraybackslash}X}
\definecolor{draftgray}{RGB}{244,245,247}
\definecolor{draftink}{RGB}{57,63,73}
\definecolor{localblue}{RGB}{50,85,128}
\definecolor{stateblue}{RGB}{230,238,248}
\definecolor{globalgreen}{RGB}{38,103,91}
\definecolor{stategreen}{RGB}{230,242,237}
\definecolor{sampleink}{HTML}{252A34}
\definecolor{sampleaccent}{HTML}{243B53}
\definecolor{samplebg}{HTML}{F1F4F8}
\definecolor{samplemuted}{HTML}{637083}
\definecolor{sampleline}{HTML}{CDD5DF}
\newenvironment{samplebox}[2]{%
  \par\addvspace{8pt}%
  \normalfont\rmfamily\fontsize{8.5}{10}\selectfont
  \enlargethispage{1.5\baselineskip}
  \color{sampleink}%
  \setlength{\parindent}{0pt}%
  \setlength{\parskip}{2pt}%
  \noindent
  \begingroup
    \setlength{\fboxsep}{8pt}%
    \colorbox{samplebg}{%
      \parbox{\dimexpr\linewidth-2\fboxsep\relax}{%
        \setlength{\parskip}{0pt}%
        \raggedright
        {\sffamily\bfseries
         \fontsize{10}{12}\selectfont
         \color{sampleaccent}#1\par}%
        \vspace{3pt}%
        {\sffamily
         \fontsize{8}{10}\selectfont
         \color{samplemuted}#2\par}%
      }%
    }%
  \endgroup
  \par\nobreak\vspace{6pt}\nobreak
  \ignorespaces
}{%
  \par\nobreak\vspace{3pt}\nobreak
  \noindent
  {\color{sampleline}\rule{\linewidth}{0.4pt}}\par
  \addvspace{8pt}%
  \ignorespacesafterend
}
\newcommand{\cmark}{\ensuremath{\checkmark}}

\usepackage{tcolorbox}
\tcbuselibrary{breakable}
\definecolor{DemoNavy}{HTML}{243B53}
\definecolor{DemoPaper}{HTML}{F7F9FC}
\definecolor{DemoBorder}{HTML}{D6DEE8}
\definecolor{DemoInk}{HTML}{253245}
\definecolor{DemoMuted}{HTML}{6B7789}
\newtcolorbox{generatedsample}[1][Generated Text]{
  breakable,
  title={#1},
  colback=DemoPaper,
  colframe=DemoBorder,
  coltext=DemoInk,
  colbacktitle=DemoNavy,
  coltitle=white,
  fonttitle=\sffamily\bfseries\small,
  fontupper=\rmfamily\fontsize{9}{12}\selectfont,
  boxrule=0.5pt,
  arc=2mm,
  outer arc=2mm,
  left=11pt,
  right=11pt,
  top=8pt,
  bottom=9pt,
  toptitle=6pt,
  bottomtitle=6pt,
  before skip=12pt,
  after skip=12pt,
  before upper={
    \setlength{\parindent}{0pt}
    \setlength{\parskip}{6pt}
  }
}

\DeclareFontShape{T1}{cmss}{bx}{sc}{<->ssub*cmss/bx/n}{}

\title{Designing Transformers for Continuous Diffusion Language Models}
\papershorttitle{Designing Transformers for Continuous Diffusion Language Models}

\author[2]{Weijian Luo\thanks{Equal contribution.}\thanks{Project leader: \texttt{pkulwj1994@icloud.com}}}
\author[1,3]{Yujian Chen\protect\footnotemark[1]}
\author[1,4]{Hao Jiang}
\author[5]{Guang Cheng}
\author[1,4]{Tianyang Hu\thanks{Corresponding author: \href{mailto:hutianyang@cuhk.edu.cn}{\texttt{hutianyang@cuhk.edu.cn}}}}
\affil[1]{Polyadic Research Group}
\affil[2]{Dots Studio, Xiaohongshu}
\affil[3]{Peking University}
\affil[4]{SAGA Lab, The Chinese University of Hong Kong, Shenzhen}
\affil[5]{The University of California, Los Angeles}
\paperpdfauthor{Weijian Luo, Yujian Chen, Hao Jiang, Guang Cheng, Tianyang Hu}

\renewcommand{\headerfont}{\normalfont\color{black}\fontsize{8}{10}\selectfont}

\begin{document}
\maketitle
\begin{abstract}
Continuous diffusion language models (CDLMs) offer an alternative approach to parallel language generation; however, most existing studies directly use transformer architectures from either causal large language models (LLMs) or Diffusion Transformers (DiTs) for image generation, which were not optimized for CDLMs. In this paper, we conduct a systematic study on Transformer architectures for CDLMs and introduce \textbf{Language Diffusion Transformer} (\textcolor{red}{\textbf{\ours}}), combining feature and query--key normalization, wider noise conditioning, likelihood-shortcut gradient routing and calibration, and confidence-gated training self-conditioning. \ours significantly reduces generation perplexity in matched LangFlow comparisons, establishing a new state of the art in continuous language generation across different model formulations. 
Using LangFlow as a representative CDLM, \ours as a plug-and-play backbone leads to a new state-of-the-art model with a generation perplexity (GenPPL) of \textcolor{red}{\textbf{21.72}} using 1024 NFEs and \textcolor{red}{\textbf{44.72}} GenPPL using 128 NFEs on the unconditional generation of 1024-token sequences on the OpenWebText dataset. When adapted to Plaid-style CDLM formulations, \ours results in a new leading GenPPL of \textcolor{red}{\textbf{11.33}} using 1024 denoising NFEs, and \textcolor{red}{\textbf{27.51}} GenPPL using 128 denoising NFEs. These gains are achieved while retaining each model’s underlying representation, training objective, and inference sampler, isolating transformer architecture as a fundamental source of improvement. Our study identifies architectural design as an important source of improvement in continuous diffusion language models.
\end{abstract}

\begin{figure}[t]
\centering
\includegraphics[width=\linewidth]{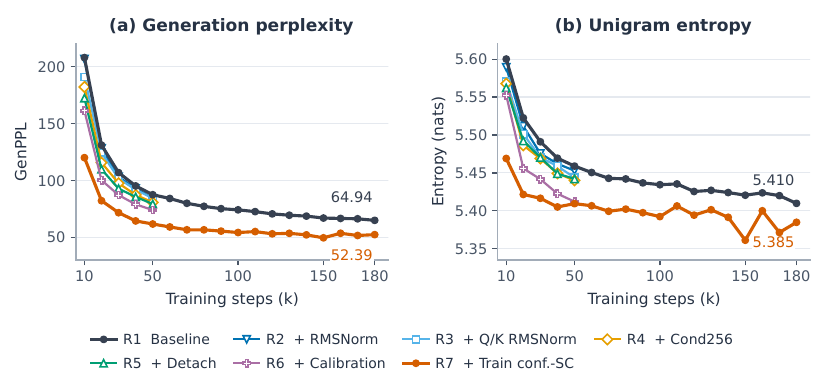}
\caption{\textbf{Architectural ablations during training on OWT-1024.} GenPPL (left) and mean per-sample unigram entropy (right) at 128 NFEs, using 1,024 samples and EMA weights. R1 (baseline) and R7 (full model) are shown through 180k updates. For clarity, trajectories for R2--R6 are shown only through 50k updates. Points are unsmoothed logged measurements; endpoints are labeled at 180k. Legend additions are cumulative, with wider conditioning preceding shortcut detachment (Table~\ref{tab:ablation}). Lower GenPPL is accompanied by lower unigram entropy; checkpoint steps and sample counts are stated separately for each comparison.}
\label{fig:tb-nfe128-curves}
\end{figure}

\section{Introduction}
\label{sec:introduction}
Autoregressive language models generate text by predicting tokens sequentially. Continuous diffusion language models (CDLMs) offer an alternative: they jointly refine continuous representations of a token sequence through iterative denoising or flow dynamics, allowing multiple token positions to be updated in parallel. Recent studies have explored a range of representations, training objectives, and generative dynamics, including one-hot flows, learned token embeddings, and latent-space models \citep{lee2026flow,yang2026replaid,chen2026langflow,hu2026elf,meshchaninov2026ldlm,guo2026cola,liang2026aurora}. Despite their different formulations, these models share a central computational component: a neural backbone that maps noisy continuous states to predictions for language generation.

\begin{figure}[!b]
\centering
\includegraphics[width=1\linewidth]{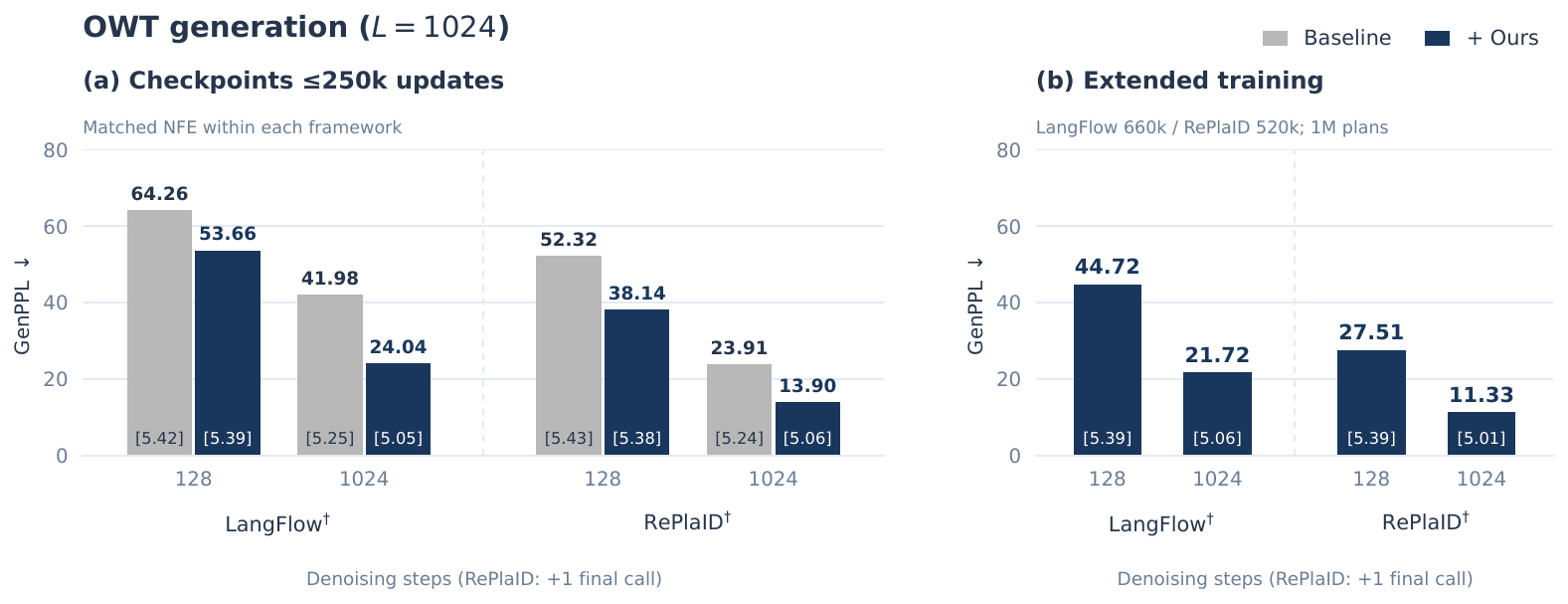}
\caption{\textbf{Generation results on OWT ($L=1024$).}
(a) Baseline and \ours{} at 128/1,024 denoising steps using checkpoints within 250k updates. The LangFlow comparisons use 250k checkpoints at 128 NFE and 180k checkpoints at 1,024 NFE. RePlaID compares 40k versus 150k at 128 steps and 220k versus 250k at 1,024 steps, so these pairs do not match training progress; the latter also differs in planned budget.
(b) Extended-training checkpoints: LangFlow 660k and RePlaID 520k from 1M-update plans. RePlaID requires one final denoiser call beyond the displayed step count.
Bars show GenPPL; bracketed values show mean per-sample unigram entropy. All results use 1,024 samples and EMA weights; $\dagger$ denotes our implementation (Table~\ref{tab:main-results}).}
\label{fig:owt-teaser}
\end{figure}

Much of the progress in CDLMs has focused on the formulation and learning of continuous generative processes, while the Transformer architectures that implement them have received comparatively less systematic attention. Existing models commonly inherit architectural choices from causal large language models (LLMs) or from Diffusion Transformers (DiTs) developed for image generation. These architectures provide useful starting points for methodological research, but their suitability for continuous language generation warrants closer examination. Causal LLMs are designed for next-token prediction, whereas image DiTs operate on continuous visual representations. CDLMs must recover discrete language tokens from noisy continuous states while coordinating noise-dependent computation, vocabulary prediction, and, when used, self-conditioning. These requirements motivate a central question:
\begin{itemize}
    \item \textbf{\emph{What Transformer architecture is best suited to continuous diffusion language models?}}
\end{itemize}

In this paper, we systematically study Transformer architectures for CDLMs and introduce \ours{}, a backbone designed for continuous language generation. Our central finding is that architectural choices effective in causal LLMs or image diffusion models do not necessarily yield the best performance in CDLMs. By varying the architecture within fixed model formulations, we show that architectural design can substantially improve generation performance. We first develop and evaluate \ours{} within LangFlow \citep{chen2026langflow}, then transfer it to Plaid-style CDLMs \citep{gulrajani2023plaid,yang2026replaid}. In each setting, we retain the underlying representation, objective formulation, and inference sampler to study architectural changes within each formulation. The Plaid adaptation reuses a subset of the design while retaining its native prediction interface.

We investigate three aspects of CDLM architecture. Within the backbone, we study how feature normalization and noise conditioning affect denoising. At the vocabulary interface, we examine how gradients through the likelihood shortcut affect the learning of token embeddings. For self-conditioning, we investigate how strongly previous predictions should guide subsequent denoising. A cumulative ablation study under shared training and evaluation settings measures the incremental effects of these components, leading to a reusable architecture that improves generation while preserving the underlying model formulation.

We evaluate \ours{} on OpenWebText unconditional generation at a sequence length of 1,024. A seven-stage LangFlow ladder tracks successive architectural changes (Figure~\ref{fig:tb-nfe128-curves}). At 250k updates and 128 NFE under the same schedule, GenPPL decreases from 64.26 to 53.66, while mean per-sample unigram entropy changes from 5.416 to 5.391. Separately, a 660k LangFlow checkpoint and a 520k Plaid checkpoint from 1M-update plans reach GenPPL 21.72 and 11.33 at the larger sampling budget (Table~\ref{tab:main-results}). These extended-training scores accompany larger changes in entropy and do not isolate the effect of architecture from training budget. We distinguish these settings throughout the paper (Figure~\ref{fig:owt-teaser}).

Our contributions are threefold: \begin{itemize} 
\item We conduct a systematic empirical study of six denoiser-design interventions in CDLMs, using a cumulative LangFlow ablation under shared training and evaluation settings. 
\item We introduce \ours{}, which combines backbone normalization and noise conditioning with likelihood-shortcut gradient routing, calibration, and confidence-gated training self-conditioning. 
\item We demonstrate lower GenPPL in matched LangFlow comparisons and examine the reuse of compatible components in a Plaid-style formulation, reporting sampling cost and unigram entropy alongside generation perplexity. \end{itemize}

Together, these findings establish architectural design as an important source of progress in CDLMs and provide a reusable foundation for further research on parallel language generation.

\section{Related Work}
\label{sec:related}

\paragraph{Continuous language diffusion and flow.}
Diffusion-LM and self-conditioned embedding diffusion established continuous denoising approaches to language generation \citep{li2022diffusionlm,strudel2022selfconditioned}, while CDCD and Plaid developed categorical diffusion and likelihood-based training \citep{dieleman2022cdcd,gulrajani2023plaid}. Recent work explores flow maps \citep{lee2026flow}, learned embeddings with information-based noise scheduling \citep{chen2026langflow}, architectural refinements \citep{yang2026replaid}, hyperspherical dynamics \citep{deschenaux2026hyperspherical}, and embedding-space flow matching \citep{hu2026elf}. We study architectural design within fixed LangFlow and Plaid-style formulations.

\paragraph{Latent representations and decoding.}
CoDAR uses contextual autoregressive decoding \citep{shen2026codar}, while LDLM jointly learns an encoder, denoiser, and decoder with controlled gradient interactions \citep{meshchaninov2026ldlm}. Cola-DLM introduces a text VAE and latent prior \citep{guo2026cola}, and AURORA-LM restricts the noisy-input pathway while preserving high-capacity latents \citep{liang2026aurora}. Analysis of ELF links token recovery to decoder margins and denoising trajectories \citep{du2026decoder}. Our changes operate within the existing codebook-based prediction interface.

\paragraph{Normalization and conditioning.}
DiT and EDM2 emphasize adaptive conditioning and magnitude control in diffusion networks \citep{peebles2023dit,karras2024edm2}. RMSNorm, nGPT, and query--key normalization provide complementary approaches to controlling representation and attention scales \citep{zhang2019rmsnorm,loshchilov2025ngpt,henry2020qknorm}. Related work examines attention-entropy collapse \citep{zhai2023entropy} and uses query--key RMSNorm in rectified-flow Transformers \citep{esser2024rectified}. We examine these normalization choices alongside noise-conditioning capacity in continuous language denoising.

\paragraph{Self-conditioning.}
Analog Bits, self-conditioned embedding diffusion, and LangFlow use previous predictions to guide denoising \citep{chen2022analogbits,strudel2022selfconditioned,chen2026langflow}. AURORA-LM addresses training--inference mismatch through trajectory consistency \citep{liang2026aurora}. Our entropy-based gate regulates bootstrap inputs during training while retaining the baseline loss formulation and inference sampler.

\paragraph{Discrete diffusion and evaluation.}
MDLM, LLaDA, and iLLaDA advance masked diffusion language modeling \citep{sahoo2024mdlm,nie2025llada,nie2026illada}, while Duo connects uniform-state discrete diffusion to an underlying Gaussian process \citep{sahoo2025duo}. Across diffusion families, likelihood rankings can differ from those based on generation quality and sampling cost \citep{sahoo2026scaling}; lower GenPPL can also accompany reduced distributional coverage \citep{franca2026hacking}. We therefore report GenPPL alongside mean per-sample unigram entropy, with explicit sampling budgets.


\section{The Language Diffusion Transformer}
\label{sec:method}

\ours{} is a Transformer denoiser for continuous language generation. Its design addresses three components of denoising: processing noisy embeddings, predicting discrete tokens, and incorporating previous predictions through self-conditioning. We develop the architecture within a LangFlow-style baseline \citep{chen2026langflow}, retaining its continuous representation, training and noise-schedule objective formulations, and inference sampler across architectural variants. Table~\ref{tab:principles} summarizes the design choices.

\begin{table}[t]
\centering
\small
\caption{Architectural design principles and their implementation in \ours{}.}
\label{tab:principles}
\begin{tabularx}{\linewidth}{
    @{}p{0.20\linewidth}
    >{\raggedright\arraybackslash}X
    >{\raggedright\arraybackslash}X@{}
}
\toprule
Component & Design principle & Implementation \\
\midrule
Backbone
& Separate feature-scale control from noise-dependent modulation.
& Backbone and Q/K RMSNorm; wider noise conditioning. \\

Vocabulary head
& Control the shortcut's forward contribution and direct codebook gradients separately.
& Shortcut detachment; bounded learned calibration. \\

Self-conditioning
& Distinguish clean-state estimation from auxiliary feedback during training.
& Confidence-gated bootstrap inputs; unweighted posterior-mean estimates at inference. \\
\bottomrule
\end{tabularx}
\end{table}

\subsection{Overall Denoising Pipeline}
\label{sec:overview}

\begin{figure}[t]
\centering
\includegraphics[width=0.95\linewidth]{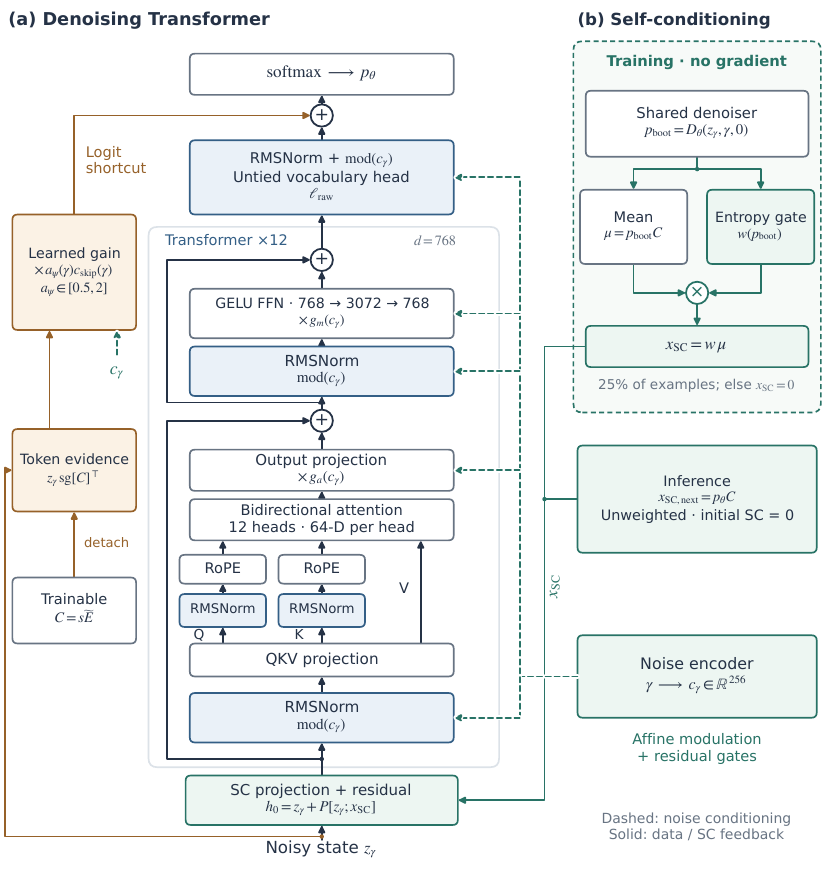}
\caption{\textbf{Transformer architecture and self-conditioning in \ours{}.}
\textbf{(a)} The denoiser is read from bottom to top, with one of 12 blocks expanded. RMSNorm precedes each modulated sublayer; headwise query--key RMSNorm precedes RoPE, while values undergo neither operation. The 256-dimensional noise embedding supplies modulation parameters and residual gates (dashed arrows). A gated likelihood shortcut bypasses the Transformer and its untied vocabulary head. Only the shortcut's explicit codebook factor is detached; $C=s\widetilde E$ remains trainable through noisy-state construction.
\textbf{(b)} A bootstrap call to the same denoiser produces token probabilities without gradient tracking. These probabilities determine a clean-state estimate and an entropy-based weight for training self-conditioning (Equation~\eqref{eq:architecture-training-sc}). Inference passes the unweighted estimate to the next step. Shortcut warmup is omitted for clarity.}
\label{fig:final-architecture}
\end{figure}

At each denoising step, an input projection combines noisy token embeddings $z_\gamma$ with an optional self-conditioning estimate $x_{\mathrm{SC}}$. A bidirectional Transformer processes the sequence through attention and feed-forward sublayers modulated by the noise level $\gamma$. An untied vocabulary head produces contextual logits, augmented by a gated likelihood shortcut that measures compatibility between the noisy state and the codebook. Softmax yields token probabilities, whose weighted embedding mean provides the clean-state estimate used for both sampler updates and subsequent self-conditioning. Figure~\ref{fig:final-architecture}(a) illustrates this pipeline with one Transformer block expanded.

Relative to the baseline, \ours{} introduces backbone RMSNorm for feature-scale control, query--key RMSNorm for attention-logit scale control, and wider noise conditioning for greater modulation capacity. The likelihood shortcut follows a Gaussian noise model \citep{gulrajani2023plaid,yang2026replaid,chen2026langflow}. We detach its embedding matrix to block the direct codebook-gradient path, while a bounded gate controls its forward contribution. An entropy-based gate further regulates bootstrap self-conditioning during training (Figure~\ref{fig:final-architecture}(b)); inference retains the un-gated posterior-mean estimator for both self-conditioning and sampler updates.

We next summarize the LangFlow formulation underlying our architectural study.

\subsection{Continuous Representation and Training Objective}
\label{sec:formulation}

LangFlow represents tokens with learned embeddings and generates language by transporting Gaussian noise toward these embeddings \citep{chen2026langflow}. Its denoiser predicts token probabilities, whose embedding mean supplies the continuous estimate required by the flow.

\paragraph{Clean embeddings and noisy inputs.}
Let $\V$ denote the vocabulary, $y_{1:L}\in\V^L$ a token sequence, and $E\in\mathbb R^{|\V|\times d}$ a learned codebook. Each row is normalized to norm $\sqrt d$ and scaled by $s=0.1$:
\begin{equation}
    \widetilde E_v=\sqrt d\,\frac{E_v}{\lVert E_v\rVert_2},
    \qquad
    C_v=s\widetilde E_v,
    \qquad v\in\V.
    \label{eq:codebook}
\end{equation}
The resulting clean-embedding table is $C=s\widetilde E$. A token sequence is represented by $x_0\in\mathbb R^{L\times d}$, with $x_{0,i}=C_{y_i}$.

For the log noise-to-signal ratio $\gamma=\log(\sigma_\gamma^2/\alpha_\gamma^2)$, we define
\begin{equation}
    \begin{gathered}
        \alpha_\gamma=\sqrt{\operatorname{sigmoid}(-\gamma)},
        \qquad
        \sigma_\gamma=\sqrt{\operatorname{sigmoid}(\gamma)},\\
        z_\gamma=\alpha_\gamma x_0+\sigma_\gamma\epsilon,
        \qquad
        \epsilon_{ij}\overset{\mathrm{i.i.d.}}{\sim}\mathcal N(0,1).
    \end{gathered}
    \label{eq:forward}
\end{equation}
Noise is independent across positions and embedding dimensions. Larger $\gamma$ corresponds to greater corruption: $\gamma\to+\infty$ yields Gaussian noise, whereas $\gamma\to-\infty$ recovers the clean state.

\paragraph{Token prediction objective.}
The denoiser predicts $p_{\theta,i}(v\mid z_\gamma,\gamma,x_{\mathrm{SC}})$, the probability of token $v$ at position $i$ given the full noisy sequence. We use $\theta$ collectively for the trainable denoiser parameters, including the codebook and, where applicable, the shortcut-gate parameters $\psi$. The noise-schedule parameters are represented separately through $\pi$. We suppress $x_{\mathrm{SC}}$ when it is not needed to distinguish the training and sampling procedures.

Training uses token-level cross-entropy:
\begin{equation}
    \mathcal L_{\mathrm{CE}}
    =
    \E_{y_{1:L}\sim p_{\mathrm{data}},\,\gamma\sim\pi,\,\epsilon}
    \left[
        -\frac{1}{L}\sum_{i=1}^{L}
        \log p_{\theta,i}(y_i\mid z_\gamma,\gamma)
    \right].
    \label{eq:langflow-ce}
\end{equation}
Here, $p_{\mathrm{data}}$ is the training sequence distribution, and $\pi$ is the learned Gumbel distribution over noise levels used by LangFlow's information-based scheduler \citep{chen2026langflow}. All variants retain this loss formulation and the baseline entropy-matching objective for the schedule.

For a fixed codebook and corruption process, the population optimum of cross-entropy recovers the token posterior when the model has sufficient capacity. The corresponding probability-weighted embedding mean therefore equals the conditional clean-state mean $\E[x_0\mid z_\gamma,\gamma]$. LangFlow formalizes this connection through Bregman-divergence flow matching \citep{chen2026langflow}, explaining how categorical prediction supports denoising and sampling in continuous space.

\subsection{Noise-Conditioned Transformer Backbone}
\label{sec:architecture-backbone}
\label{sec:normalization}
\label{sec:structure}
\label{sec:conditioning}

The backbone processes noisy embeddings while adapting to the noise level. We study normalization and noise conditioning as separate components: normalization controls feature scales, while conditioning supplies noise-dependent modulation.

\paragraph{Input and self-conditioning fusion.}
The input state is
\begin{equation}
    h_0=z_\gamma+P([z_\gamma;x_{\mathrm{SC}}]),
    \label{eq:architecture-input}
\end{equation}
where brackets denote concatenation along the feature dimension and $P:\mathbb R^{2d}\to\mathbb R^d$ is a zero-initialized, bias-free linear projection applied independently at each position. The embedding and backbone widths are equal in this architecture. When self-conditioning is absent, $x_{\mathrm{SC}}=0$.

\paragraph{Transformer blocks.}
Each block contains bidirectional attention and a feed-forward sublayer, each with a noise-conditioned residual gate. Following DiT \citep{peebles2023dit}, a noise embedding $c_\gamma$ supplies the shifts, scales, and gates:
\begin{align}
    u &= h+g_a(c_\gamma)\odot
    \operatorname{Attn}\!\left(
        \operatorname{mod}_a(N_a(h);c_\gamma)
    \right), \nonumber\\
    h' &= u+g_m(c_\gamma)\odot
    \operatorname{MLP}\!\left(
        \operatorname{mod}_m(N_m(u);c_\gamma)
    \right).
    \label{eq:block}
\end{align}
The subscripts $a$ and $m$ identify the attention and MLP sublayers; block indices are omitted for readability. We write the affine modulation as
\begin{equation}
    \operatorname{mod}_j(v;c_\gamma)
    =
    \bigl(1+\delta_j(c_\gamma)\bigr)\odot v+\beta_j(c_\gamma),
    \qquad j\in\{a,m\},
    \label{eq:architecture-modulation}
\end{equation}
where modulation vectors and residual gates are broadcast across sequence positions.

\paragraph{Backbone RMSNorm.}
LayerNorm subtracts the feature mean before normalizing magnitude. To examine whether this re-centering is needed in embedding-space denoising, we replace the two pre-normalization layers in each Transformer block and the normalization layer in the vocabulary head with RMSNorm \citep{zhang2019rmsnorm}:
\begin{equation}
    \RMSNorm(x)=g\odot\frac{x}{\sqrt{d^{-1}\sum_{j=1}^{d}x_j^2+\varepsilon}},
    \qquad \varepsilon=10^{-6}.
    \label{eq:rmsnorm}
\end{equation}
The mean-square reduction is computed in FP32, and the output is cast to the projection dtype. Both normalization layers use a learned gain without a bias, so the replacement preserves the parameter count. RMSNorm rescales the input without removing its mean component. The ablation tests the effect of this change while retaining the baseline adaptive modulation.

\paragraph{Query--key RMSNorm.}
Normalizing the input to attention does not directly constrain the magnitudes produced by the learned query and key projections. In dot-product attention, changes in $\lVert q\rVert\lVert k\rVert$ alter logit scale independently of directional similarity. We therefore apply RMSNorm to each 64-dimensional query and key vector after the fused QKV projection and head reshape, before rotary position embeddings:
\begin{equation}
    \widehat q_{i,h}=\RMSNorm(q_{i,h}),\qquad
    \widehat k_{i,h}=\RMSNorm(k_{i,h}).
    \label{eq:qknorm}
\end{equation}
Queries and keys have separate learned gains, each shared across heads within a layer; values are unchanged. The design follows QK normalization \citep{henry2020qknorm} and its RMSNorm implementation in rectified-flow Transformers \citep{esser2024rectified}. For head dimension $d_h$ and gain vectors $g_q,g_k$, the normalized dot-product logit obeys
\begin{equation}
    \left|\frac{(R_i\widehat q)^\top(R_j\widehat k)}{\sqrt{d_h}}\right|
    \leq \sqrt{d_h}\,\lVert g_q\rVert_\infty\lVert g_k\rVert_\infty,
    \label{eq:attention-bound}
\end{equation}
where $R_i,R_j$ are the orthogonal rotary-position transformations. The bound removes raw input norms from the right-hand side; it does not fix a temperature or bound learned gains over training. Appendix~\ref{app:attention-analysis} gives the derivation. The ablations assess whether this scale control improves generation.

\paragraph{Noise-conditioning capacity.}
The conditioning vector supplies shifts, scales, and gates to all 12 Transformer blocks and modulation parameters to the vocabulary head. We increase its width from 128 to 256, changing the timestep MLP from $256\!\rightarrow\!128\!\rightarrow\!128$ to $256\!\rightarrow\!256\!\rightarrow\!256$. The input is a 256-dimensional sinusoidal encoding of $\gamma$, and the MLP uses SiLU after each linear layer. All adaptive-normalization projections and the shortcut gate receive the wider vector. This increases the capacity for noise-dependent modulation without changing the Transformer hidden width or feed-forward dimensions. The change adds 7.36M parameters, or 2.25\% of the baseline; its ablation therefore includes a capacity increase and is not a parameter-matched test of conditioning alone.

\paragraph{Contextual vocabulary head.}
After the Transformer blocks, an RMS-normalized, noise-modulated head maps hidden states to contextual vocabulary logits $\ell_{\mathrm{raw},\theta}\in\mathbb R^{L\times|\V|}$. Its output weights are untied from the clean-embedding table $C$. The head draws on the full noisy sequence, while the likelihood shortcut supplies direct evidence from each position.

\subsection{Vocabulary Prediction with a Likelihood Shortcut}
\label{sec:architecture-vocabulary}
\label{sec:shortcut}

Under Gaussian corruption, the distance between a noisy state $z_{\gamma,i}$ and a token's signal-scaled embedding $\alpha_\gamma C_v$ provides local evidence about that token. Because all codebook rows have the same norm, the Gaussian observation log-likelihood reduces to a scaled dot product, up to a token-independent constant.

The baseline adds this analytic likelihood shortcut to the contextual logits:
\begin{equation}
    \begin{gathered}
        \ell_\theta^{\mathrm{base}}(z_\gamma,\gamma)
        =
        \ell_{\mathrm{raw},\theta}(z_\gamma,c_\gamma)
        +c_{\mathrm{skip}}(\gamma)\,z_\gamma C^\top,\\
        c_{\mathrm{skip}}(\gamma)
        =\frac{\alpha_\gamma}{\sigma_\gamma^2}.
    \end{gathered}
    \label{eq:baseline-logits}
\end{equation}
This parameterization follows Plaid and LangFlow \citep{gulrajani2023plaid,chen2026langflow}. The equation shows the shortcut at full strength. During baseline training, its contribution is linearly warmed up over the first 5,000 optimization steps.

\ours{} modifies both the gradient path through this branch and its forward contribution:
\begin{equation}
    \ell_\theta^{\mathrm{ours}}
    =
    \ell_{\mathrm{raw},\theta}
    +a_\psi(\gamma)c_{\mathrm{skip}}(\gamma)\,
    z_\gamma\operatorname{sg}[C]^\top.
    \label{eq:architecture-logits}
\end{equation}
Here, $\operatorname{sg}$ denotes stop-gradient, and $a_\psi(\gamma)$ is a learned bounded gate computed from $c_\gamma$. This gate calibrates the likelihood shortcut and is separate from the residual gates within the Transformer.

Stop-gradient leaves the forward logits unchanged at fixed parameters and removes only the gradient through the shortcut's explicit codebook factor. The codebook remains trainable through noisy-state construction. Detachment therefore changes the optimization graph without changing the forward model class or adding parameters. Appendix~\ref{app:gradient-analysis} derives the removed contribution; this identity does not itself establish gradient conflict.

\paragraph{Bounded calibration.}
The contextual head sees the full noisy sequence and is not constrained to model a context-only prior. We therefore allow a structured adjustment of the shortcut's contribution:
\begin{equation}
    a_\psi(\gamma)=\exp\!\left(
    \operatorname{clip}(f_\psi(c_\gamma),-\log 2,\log 2)\right).
    \label{eq:calibrated-shortcut}
\end{equation}
Here $f_\psi$ is a $256\!\rightarrow\!32\!\rightarrow\!1$ MLP with SiLU activation and a zero-initialized final layer. Thus $a_\psi\in[0.5,2]$ and initially equals one. Both baseline and modified models retain the 5,000-step linear shortcut warmup. Calibration changes the forward weighting, while detachment controls the direct codebook gradient. An unrestricted head could absorb a change in shortcut weighting, so this parameterization is not a correction to the Gaussian likelihood identity (Appendix~\ref{app:observation-analysis}).

\subsection{Self-Conditioning during Training and Sampling}
\label{sec:architecture-self-conditioning}

Self-conditioning supplies a clean-state estimate as an auxiliary denoiser input. \ours{} changes how this input is weighted during training while preserving the estimate used by the inference sampler.

\paragraph{Training with gated self-conditioning.}
Following the baseline, we use self-conditioning on 25\% of training examples, adopting the bootstrap procedure used in prior work \citep{chen2022analogbits,strudel2022selfconditioned}. For these examples, the full denoiser is first evaluated at the current $z_\gamma$ and $\gamma$, with self-conditioning set to zero. This bootstrap call runs without gradient tracking and produces a probability matrix $p_{\mathrm{boot}}\in\mathbb R^{L\times|\V|}$.

The bootstrap probabilities determine a clean-state estimate and its confidence weight:
\begin{equation}
    \mu_{\mathrm{boot}}=p_{\mathrm{boot}}C,
    \qquad
    x_{\mathrm{SC}}^{\mathrm{train}}
    =
    \operatorname{sg}\!\left[w(p_{\mathrm{boot}})\odot\mu_{\mathrm{boot}}\right],
    \label{eq:architecture-training-sc}
\end{equation}
where $w(p_{\mathrm{boot}})\in\mathbb R^{L\times 1}$ is broadcast across embedding dimensions. The weight at position $i$ is computed from the token entropy
\begin{equation}
    H_i(p_{\mathrm{boot}})
    =
    -\sum_{v\in\V}
    p_{\mathrm{boot},i}(v)\log p_{\mathrm{boot},i}(v).
    \label{eq:architecture-bootstrap-entropy}
\end{equation}

The token-wise weight is
\begin{equation}
    w_i(p)=\operatorname{clip}\!\left(1-\frac{H_i(p)}{\log|\V|},0,1\right).
    \label{eq:confidence-sc}
\end{equation}
It is zero for a uniform distribution and one for a one-hot distribution. The complete bootstrap input, including multiplication by the codebook, is computed without gradients. The main denoiser call receives the weighted estimate. The gate therefore controls the influence of the auxiliary input without changing the definition of $\mu_{\mathrm{boot}}$. The bootstrap prediction is derived from the current training input; it does not come from an earlier sampling step. The remaining 75\% of examples use zero self-conditioning.

\paragraph{Interpretation.} We use token entropy as a heuristic measure of bootstrap confidence. Our hypothesis is that attenuating diffuse bootstrap predictions during training reduces the denoiser's reliance on unreliable auxiliary inputs. The gate changes the training feedback signal, while both the inference sampler and inference-time self-conditioning retain the unweighted posterior mean. Appendix~A.4 distinguishes token entropy from uncertainty in embedding space and explains why confidence weighting should not be interpreted as an improved clean-state estimator.

\paragraph{Sampling with unweighted estimates.}
Generation starts from Gaussian noise and proceeds from high to low $\gamma$ using the baseline Euler--EDM sampler \citep{karras2022edm}. The normalized corruption coordinates in Equation~\eqref{eq:forward} correspond to the additive-noise coordinates
\begin{equation}
    \tau_\gamma
    =\frac{\sigma_\gamma}{\alpha_\gamma}
    =e^{\gamma/2},
    \qquad
    \bar z_\gamma
    =\frac{z_\gamma}{\alpha_\gamma}
    =x_0+\tau_\gamma\epsilon,
    \label{eq:architecture-edm-coordinates}
\end{equation}
where $\tau_\gamma$ is the EDM noise scale. The denoiser receives the normalized state $z_\gamma=\alpha_\gamma\bar z_\gamma$.

At each non-final step, softmax applied to the combined logits yields token probabilities and the clean-state estimate
\begin{equation}
    \widehat x_{0,i}
    =
    \sum_{v\in\V}
    p_{\theta,i}(v\mid z_\gamma,\gamma,x_{\mathrm{SC}})\,C_v.
    \label{eq:clean-prediction}
\end{equation}
This unweighted estimate determines the solver update and supplies self-conditioning for the next denoising step. The first step uses zero self-conditioning. Sampling requires neither an additional bootstrap call nor entropy-based weighting, including for models trained with confidence-gated self-conditioning. The final prediction is decoded by token-wise argmax.

\subsection{Adaptation to a Plaid-Style Formulation}
\label{sec:transfer}

Our Plaid-style baseline is our implementation of RePlaID \citep{yang2026replaid}: it combines a 16-dimensional spherical codebook, a likelihood-bound objective (NELBO), a learned noise schedule, and ancestral DDPM sampling with a 12-block, width-768 Transformer. It is not the original published Plaid checkpoint. Unlike LangFlow, its continuous-state width is 16 rather than 768, and its native projections map the noisy state and self-conditioning into the Transformer. Thus Equations~\eqref{eq:codebook} and~\eqref{eq:architecture-input} specify the LangFlow instantiation, not a replacement of Plaid's representation or input interface.

\begin{table}[t]
\centering\small
\caption{Scope of the Plaid adaptation. Four of the six LangFlow interventions are transferred; the native likelihood interface and gradient rules are retained.}
\label{tab:transfer}
\begin{tabularx}{\linewidth}{@{}lYY@{}}
\toprule
Component & Plaid baseline & Plaid + \ours{} \\
\midrule
Backbone / head norm & LayerNorm & RMSNorm \\
Query--key norm & None & Headwise RMSNorm \\
Condition width & 128 & 256 \\
Shortcut gradients & Codebook factor detached on self-conditioned training examples & Same native rule \\
Shortcut calibration & Analytic coefficient & Same; no learned multiplier \\
Training self-conditioning & Unweighted bootstrap mean & Entropy-weighted bootstrap mean \\
\bottomrule
\end{tabularx}
\end{table}

The native shortcut already has selective codebook detachment on self-conditioned examples. The adapted model retains that rule rather than importing LangFlow's unconditional branch detachment, and does not add the learned calibration gate. Both variants retain the 16-dimensional state, NELBO, learned schedule, and DDPM transitions. The reported adapted-model evaluations explicitly select unweighted inference self-conditioning; enabling its optional inference gate would be a separate sampling intervention. The extended run uses a 1M-update plan, with a resumed decay schedule and an evaluated 520k checkpoint; Appendix~\ref{app:extended-training} gives the optimization settings. This experiment assesses reuse of compatible components, not transfer of the entire LangFlow pipeline.

\subsection{Implementation Details}
\label{sec:architecture-implementation}

For OpenWebText, \ours{} uses 12 Transformer blocks with embedding and hidden width $d=768$, 12 attention heads of dimension 64, and feed-forward width 3072. For the LangFlow formulation, the input projection $P$ maps 1536 dimensions to 768 dimensions. Transformer depth and hidden width are retained from \base{}.

Both the baseline and \ours{} use fused QKV projections, rotary position embeddings, GELU feed-forward layers, and zero-initialized adaptive-normalization projections. Configuration definitions and training settings are provided in Appendix~\ref{app:runs} and~\ref{app:config}.

\subsection{Architecture Variants}
\label{sec:variants}

We evaluate the complete design through seven cumulative configurations: baseline LangFlow, followed by backbone RMSNorm, Q/K RMSNorm, increased conditioning width, shortcut detachment, learned calibration, and training-time confidence gating. Each row of Table~\ref{tab:ablation} adds one modification to the preceding configuration, with a separate training run from scratch. The final row defines \ours{}, the architecture in Figure~\ref{fig:final-architecture}, with about 100M transformer parameters including the codebook and untied vocabulary projection. The cumulative ladder uses a 250k-update schedule; extended-training runs are reported separately.

\begin{generatedsample}[Text Generation Sample of RePlaID + \ours]
{\sffamily\fontsize{8}{10}\selectfont\color{DemoMuted}
part of 1024-length sequence, 1024 Denoising NFEs\enspace\textbar\enspace GenPPL \textbf{11.33}\par}

... Through our continued friendship and with us as a company, let it be
known that we have another season to go for these talented individuals
to reach their full potential and return to the table to meet fans
expectations. We all know that this is a very exciting time to come for
us at UW{-}Meison University and we are very, very excited to see an
opportunity at our new organization. While representing our country,
an exciting time has passed since we challenged our women's to victory
in the first round of the championship event. Furthermore, we have
another season to go. While representing our country, there is no doubt
that many of our fans will open their eyes towards our debut in the
T{-}Mobile Arena at WrestleMania (sticking to win the title) and watch
with the best of our best. Furthermore, we have another season to go,
for our individuals to reach their full potential and return to the
table to meet fans expectations. We all know that this is a very
exciting time to come for us here at UW{-}Meison University and we are
very, very excited to see an opportunity at our new organization.
While representing our country, I would like to state that there is
no doubt that many of our fans will open their eyes towards our debut
at the T{-}Mobile Arena at WrestleMania. Moreover, we have a lot of work
to do, for these individuals to reach their full potential and mature
to meet our community's expectations. We only have one season to go.
While representing our country, our eyes will surely be open to the
hard work and dedication we all continue to see from the air. ...
\end{generatedsample}
\section{Experiments}
\label{sec:empirical}

We distinguish a cumulative LangFlow ablation under shared training settings from evaluations of extended-training checkpoints and the Plaid-style adaptation. Checkpoint progress, planned training budget, sampling cost, and sample count are reported separately.

\subsection{Experimental Setup}
\label{sec:settings}

\paragraph{Data and optimization.}
We train from scratch on OpenWebText using Qwen3 tokenization (151,936 output entries), document BOS/EOS markers, packing, a 10,000-document shuffle buffer, and random crops of 1,024 tokens. The LangFlow ablation uses a global batch of 512, AdamW with weight decay 0.01, gradient clipping at 1.0, and a 250k-update schedule: 5k warmup steps to $3\times10^{-4}$ followed by cosine decay to $3\times10^{-5}$. All seven configurations share the data order, objective, schedule, and sampler. Training uses bfloat16, dropout 0.1, seed 42, and EMA decay 0.999. The separate extended-training schedules are described in Appendix~\ref{app:extended-training}.

\paragraph{Generation and sampling cost.}
All local results reported here use 1,024 unconditional samples of 1,024 model tokens, temperature 1, and EMA weights. Generation seeds are 0--1,023. LangFlow uses the Euler--EDM sampler with unweighted self-conditioning and final argmax decoding. Its NFE counts all denoiser calls. Plaid uses ancestral DDPM: $K$ reverse transitions require $K+1$ model forwards, including the final denoiser call. We show this extra call explicitly. These conventions also apply to the qualitative examples in Appendix~\ref{app:qualitative}.

\paragraph{Metrics.}
We score decoded text with a frozen GPT-2 Large model. The evaluator uses GPT-2 tokenization, adds no special tokens, excludes padding, and truncates each sample to at most 1,024 evaluator tokens. Generator special tokens are removed in the clean-text field; the inline scorer falls back to raw decoded text if that field is empty. For tokenized sample $x$ of length $T_x$, corpus-level generation perplexity is
\begin{equation}
    \operatorname{GenPPL}_r(\mathcal S)=
    \exp\!\left(-\frac{\sum_{x\in\mathcal S}\sum_{t=2}^{T_x}
    \log r(x_t\mid x_{<t})}{\sum_{x\in\mathcal S}(T_x-1)_+}\right),
    \label{eq:genppl}
\end{equation}
where $(a)_+=\max(a,0)$; the first token has no preceding context and is not scored. Entropy denotes \emph{mean per-sample unigram entropy in nats}, computed on the same truncated, non-padding GPT-2 token sequences, including their first tokens. Samples are weighted equally, not pooled into a corpus histogram (Appendix~\ref{app:reporting}). This lexical statistic does not establish semantic diversity or distributional coverage, so a lower GenPPL is not by itself evidence of better overall generation quality.

\subsection{Generation Results}
\label{sec:main-results}

\paragraph{Comparison with public models.}
Table~\ref{tab:main-results} combines published autoregressive and diffusion results with our local LangFlow and Plaid-style results. Published OWT GenPPL includes 104.9 for MDLM, 103.6 for SEDD Uniform, 77.6 for Duo, 62.2 for FLM, 36.5 for LangFlow, and 35.9 for the autoregressive Transformer. The local rows identify actual checkpoint updates and denoiser calls, with checkpoints within 250k updates and extended-training results grouped separately.

\begin{table}[t]
\centering
\small
\setlength{\tabcolsep}{3pt}
\caption{Generation results on OWT ($L=1024$).
Published baseline results are taken from Table~2 of
\citet{chen2026langflow} and retain their original protocols;
local results use our implementations, 1,024 samples, and EMA weights.
Checkpoint denotes completed optimizer updates, and NFE counts
denoiser evaluations. Extended-training checkpoints come from
1M-update training plans.
Entropy denotes mean per-sample unigram entropy
(Section~\ref{sec:settings}).
A dash indicates an unreported value;
$\dagger$ denotes our implementation.
For RePlaID, $K+1$ counts $K$ reverse transitions and
the final denoiser call.}
\label{tab:main-results}
\label{tab:published-results}
\begin{tabular*}{\linewidth}{
    @{\extracolsep{\fill}}lrrrr@{}
}
\toprule
& & & \multicolumn{2}{c}{OWT ($L=1024$)} \\
\cmidrule(l){4-5}
Model & Train Steps & NFE & GenPPL$\downarrow$ & Entropy \\
\midrule

\multicolumn{5}{@{}l}{
    \emph{Published results: Autoregressive}
} \\
Transformer
    & -- & -- & 35.9 & -- \\

\midrule
\multicolumn{5}{@{}l}{
    \emph{Published results: Absorbing-state diffusion}
} \\
D3PM Absorb
    & -- & -- & -- & -- \\
DiffusionBert
    & -- & -- & -- & -- \\
SEDD Absorb
    & -- & -- & -- & -- \\
MDLM
    & -- & -- & 104.9 & -- \\
\midrule
\multicolumn{5}{@{}l}{
    \emph{Published results: Uniform-state / continuous diffusion}
} \\
D3PM Uniform
    & -- & -- & -- & -- \\
Diffusion-LM
    & -- & -- & -- & -- \\
RePlaID
    & -- & -- & -- & -- \\
SEDD Uniform
    & -- & -- & 103.6 & -- \\
UDLM
    & -- & -- & -- & -- \\
Duo
    & -- & -- & 77.6 & -- \\
FLM (1024 steps)
    & -- & -- & 62.2 & -- \\
LangFlow
    & -- & -- & \textbf{36.5} & -- \\
\midrule
\multicolumn{5}{@{}l}{
    \emph{Local results: Within 250k updates}
} \\
LangFlow$\dagger$
    & $\leq$250k & $128$ & 64.26 & 5.42 \\
\textbf{LangFlow$\dagger$ + \ours{}}
    & $\leq$250k & $128$ & 53.66 & 5.39 \\
\addlinespace[2pt]
LangFlow$\dagger$
    & $\leq$250k & $1024$ & 41.98 & 5.25 \\
\textbf{LangFlow$\dagger$ + \ours{}}
    & $\leq$250k & $1024$ & 24.04 & 5.05 \\
LangFlow$\dagger$
    & $\leq$250k & $1024$ & 39.40 & 5.22 \\
\addlinespace[2pt]
RePlaID$\dagger$
    & $\leq$250k & $128+1$ & 52.32 & 5.43 \\
\textbf{RePlaID$\dagger$ + \ours{}}
    & $\leq$250k & $128+1$ & 38.14 & 5.38 \\
\addlinespace[2pt]
RePlaID$\dagger$
    & $\leq$250k & $1024+1$ & 23.91 & 5.24 \\
\textbf{RePlaID$\dagger$ + \ours{}}
    & $\leq$250k & $1024+1$ & \textbf{13.90} & 5.06 \\
\midrule
\multicolumn{5}{@{}l}{
    \emph{Local results: Extended training}
} \\
\textbf{LangFlow$\dagger$ + \ours{}}
    & 660k & $128$ & 44.72 & 5.39 \\
\textbf{LangFlow$\dagger$ + \ours{}}
    & 660k & $1024$ & 21.72 & 5.06 \\
\addlinespace[2pt]
\textbf{RePlaID$\dagger$ + \ours{}}
    & 520k & $128+1$ & 27.51 & 5.39 \\
\textbf{RePlaID$\dagger$ + \ours{}}
    & 520k & $1024+1$ & \textbf{11.33} & 5.01 \\
\bottomrule
\end{tabular*}
\end{table}

\paragraph{Evaluation scope.}
The public-model results retain their source's training settings and evaluation protocols. In particular, the published LangFlow OWT evaluation uses GPT-2 tokenization, a 1M-update training schedule, 1,024 sampling steps, and sample-averaged perplexity. Our local evaluations use the corpus-level aggregation in Equation~\ref{eq:genppl}. The table therefore provides a broad comparison of reported results; architectural attribution uses the shared-schedule local comparison below.

\paragraph{Matched LangFlow comparison.} Under the shared 250k-update schedule, \ours{} reduces GenPPL from 64.26 to 53.66 at 128 NFE, a relative reduction of 16.5\%. Mean per-sample unigram entropy changes from 5.416 to 5.391. The comparison holds the data order, batch size, objective formulation, and sampler fixed. The modified model has 334.45M parameters, compared with 327.09M for the baseline.

\paragraph{Additional checkpoints within 250k updates.}
At 1,024 NFE, the paired LangFlow checkpoints at 180k updates yield GenPPL 41.98 and 24.04, with entropy 5.248 and 5.049. For RePlaID at 128 denoising steps, the available baseline and modified EMA evaluations are at 40k and 150k updates (52.32 and 38.14 GenPPL). At 1,024 steps they are at 220k and 250k updates (23.91 and 13.90). These RePlaID pairs differ in training progress, and the modified 250k checkpoint comes from a 1M-update plan. Figure~\ref{fig:owt-teaser}(a) labels the actual checkpoints; these results describe performance within the update cap and do not isolate the effect of architecture under matched training.

\paragraph{Extended training and transfer.}
The 660k LangFlow checkpoint reaches GenPPL 44.72 at 128 calls and 21.72 at 1,024 calls, with entropy decreasing from 5.391 to 5.056. The Plaid adaptation at 520k reaches 27.51 and 11.33 at 129 and 1,025 total calls, with entropy decreasing from 5.393 to 5.014. Thus the lower GenPPL at the larger sampling budget accompanies a larger entropy change. Both checkpoints belong to 1M-update plans; neither result is a completed 1M-step evaluation. These checkpoints characterize performance under extended training but do not isolate architectural effects. Section~\ref{sec:transfer} specifies the components adapted to Plaid. The reported Plaid scores establish an application of the design under a different objective, but do not supply a matched baseline comparison for all six LangFlow interventions.

\subsection{Cumulative Ablation and Analysis}
\label{sec:ablation}
\label{sec:mechanisms}

Each row of Table~\ref{tab:ablation} adds one change to the preceding configuration and is trained from scratch. Comparisons at 50k updates use identical sample counts, EMA conventions, and 128-NFE sampling. The table reports total parameters, including both the codebook and the untied vocabulary projection; these are distinct from the approximately 100M parameters in the Transformer and conditioning pathways.

\begin{table}[t]
\centering
\small
\caption{Cumulative architectural ablation on OWT-1024, from R1 to R7 (the complete model). Each successive row adds one component, with wider conditioning before shortcut detachment. The 50k-step column reports unsmoothed GenPPL using 128 NFE, 1,024 generated samples, and temperature 1. The 170k-step column reports unsmoothed GenPPL using 1024 NFE, 1,024 generated samples, and temperature 1.}
\label{tab:ablation}
\label{tab:tb-nfe128-ablation-50k}
\footnotesize
\setlength{\tabcolsep}{1.5pt}
\begin{tabular*}{\linewidth}{@{\extracolsep{\fill}}lccccccrrccccc@{}}
\toprule
\multicolumn{8}{c}{Cumulative architecture} & \shortstack{50k steps\\128 NFE} & \multicolumn{5}{c}{\shortstack{170k steps \\1024 NFE}} \\
\cmidrule(lr){1-8}\cmidrule(lr){9-9}\cmidrule(l){10-14}
Run & RMS & QK & $c{=}256$ & Detach & Calib. & \shortstack{Train\\Conf.-SC} & Transformer Params & GenPPL$\downarrow$ & GenPPL$\downarrow$ & Ent. \\
\midrule
R1(baseline) & & & & & & & $\sim$100M & 87.48 & 41.97 & 5.24 & & & \\
R2 & \cmark & & & & & & $\sim$100M & 84.41 & - & -& & & \\
R3 & \cmark & \cmark & & & & & $\sim$100M & 84.37 & -& -& & & \\
R4 & \cmark & \cmark & \cmark & & & & $\sim$100M & 80.74 & -& -& & & \\
R5 & \cmark & \cmark & \cmark & \cmark & & & $\sim$100M & 79.35 & -& -& & & \\
R6 & \cmark & \cmark & \cmark & \cmark & \cmark & & $\sim$100M & 74.03 & -& -& & & \\
R7 & \cmark & \cmark & \cmark & \cmark & \cmark & \cmark & $\sim$100M & \textbf{61.73} & \textbf{28.74} & 5.15 & & & \\
\bottomrule
\end{tabular*}
\end{table}

\paragraph{Normalization and conditioning.}
Backbone RMSNorm reduces 50k-step GenPPL from 87.48 to 84.41. Adding Q/K RMSNorm changes it only slightly, to 84.37, while widening noise conditioning reduces it further to 80.74. This early checkpoint alone understates the observed separation between the two normalization configurations: at 250k, R2 and R3 reach 60.76 and 57.86. The Q/K comparison therefore depends on training progress. The conditioning result includes a 7.36M-parameter capacity increase and cannot isolate the benefit of where that capacity is allocated.

\paragraph{Shortcut and training feedback.}
On the current cumulative path, detachment reduces GenPPL from 80.74 to 79.35, and calibration reduces it to 74.03. Adding confidence-gated training self-conditioning produces the largest adjacent reduction, from 74.03 to 61.73. Overall, R1--R7 reduce GenPPL by 29.4\% at 50k. These increments are conditional on earlier modifications; they are not independent contributions or evidence of statistical significance from a single training seed. Because inference uses unweighted self-conditioning throughout this ladder, the final increment reflects a change to training feedback rather than an inference gate. Constant-attenuation and interaction controls remain unreported (Appendix~\ref{app:controls}).

\paragraph{Training progress and entropy.}
Figure~\ref{fig:tb-nfe128-curves} shows the recorded trajectories, with intermediate configurations displayed only through 50k for clarity. At 180k, the baseline and full model have GenPPL 64.94 and 52.39, with entropy 5.410 and 5.385. The paired measurements support reading the GenPPL gains together with lexical-diversity changes. They do not establish the proposed optimization mechanisms: attention diagnostics and gradient-alignment measurements are not reported.

\subsection{Parameters and Computational Cost}
\label{sec:cost}

The full LangFlow model has 334.45M total parameters, compared with 327.09M for the baseline. Both counts include the $151{,}936\times768$ codebook and the untied vocabulary-projection matrix. Excluding those two matrices leaves approximately 101.07M and 93.71M parameters, respectively, including conditioning and the remaining prediction-interface parameters. Q/K RMSNorm adds 1,536 parameters and calibration adds 8,257; the remaining increase comes from widening conditioning. Replacing backbone normalization, detaching the shortcut, and applying confidence gating add none.

Equal training updates and denoiser-call counts do not establish equal FLOPs or wall-clock time. No training-throughput or sampling-latency advantage is claimed. The fixed-ID generations in Appendix~\ref{app:qualitative} cover all twelve evaluations in Figure~\ref{fig:owt-teaser}, with multiple complete outputs per setting; they are not a matched quantitative comparison.

The results support improving CDLMs through changes to the denoiser backbone and its prediction and training interfaces. Within the cumulative LangFlow study, the largest adjacent GenPPL reduction at 50k updates comes from confidence-weighted training self-conditioning. The Plaid-style adaptation illustrates how compatible components can be reused while retaining a different continuous state space, objective, and sampler. 

The study is limited to OWT-1024 at one backbone scale and one training seed. The cumulative ablation measures effects along a fixed design sequence, and wider conditioning increases parameter count. Parameter- and compute-matched comparisons, component-removal ablations, and additional seeds would help isolate individual effects and assess their robustness. Direct measurements of attention behavior and codebook gradients are also needed to test the proposed optimization explanations. The unmatched Plaid comparisons and extended-training results should be interpreted with their respective training budgets. Finally, GenPPL and unigram entropy capture limited aspects of generation quality; evaluation of coherence, semantic diversity, and distributional coverage remains necessary.
\section{Conclusion}
\label{sec:conclusion}

We have presented \ours{}, a Transformer backbone developed through a systematic study of architectural design for continuous diffusion language models. Its design separates feature-scale control from noise modulation, decouples the likelihood shortcut's forward contribution from its direct codebook gradient, and distinguishes clean-state estimation from the weighting of self-conditioning inputs. Experiments on OWT-1024 demonstrate improved generation performance within LangFlow and an adaptation of compatible components to a Plaid-style model, while preserving each model's representation, objective formulation, and inference sampler. Evaluating GenPPL alongside unigram entropy allows these gains to be interpreted together with changes in lexical diversity. These findings highlight architecture as a reusable source of improvement in CDLMs and motivate further study at larger scales and across broader generation tasks.

\label{main-text-end}

\clearpage
\bibliography{references}
\bibliographystyle{plainnat}
\clearpage
\appendix
\section{Architectural Analysis and Derivations}
\label{app:analysis}

This appendix derives the normalization, likelihood, gradient, and estimation identities used to motivate \ours{}. The analysis characterizes the architectural interventions without establishing convergence or generation-quality guarantees. We assume real arithmetic, nonzero codebook rows, and finite noise levels with $\sigma_\gamma>0$.

We retain the codebook notation $C=s\widetilde E\in\mathbb R^{|\V|\times d}$ from Equation~\eqref{eq:codebook}, so each row has squared norm $s^2d$. Single-position embeddings are treated as column vectors: $c_v=C_{v,:}^{\top}$ denotes the embedding of token $v$, and $z_{\gamma,i}\in\mathbb R^d$ denotes the column-vector representation of the noisy embedding at position $i$. The full noisy sequence remains $z_\gamma\in\mathbb R^{L\times d}$.

\subsection{Scale Dependence of Normalized Attention}
\label{app:attention-analysis}

For a query vector $q\in\mathbb R^{d_h}$ and learned gain $g_q\in\mathbb R^{d_h}$, define
\[
    \widehat q
    =
    \frac{g_q\odot q}
    {\sqrt{\lVert q\rVert_2^2/d_h+\varepsilon}},
    \qquad \varepsilon>0.
\]
Then
\begin{equation}
    \lVert\widehat q\rVert_2^2
    \leq
    \lVert g_q\rVert_\infty^2
    \frac{\lVert q\rVert_2^2}
    {\lVert q\rVert_2^2/d_h+\varepsilon}
    \leq d_h\lVert g_q\rVert_\infty^2.
    \label{eq:normalized-query-bound}
\end{equation}
The same argument applies to a normalized key $\widehat k$ with gain $g_k$. Let $R_i$ and $R_j$ denote the orthogonal rotary transformations at query position $i$ and key position $j$. For standard scaled dot-product attention, norm preservation and Cauchy--Schwarz give
\begin{equation}
    \left|
        \frac{\langle R_i\widehat q_i,R_j\widehat k_j\rangle}
        {\sqrt{d_h}}
    \right|
    \leq
    \sqrt{d_h}\,
    \lVert g_q\rVert_\infty
    \lVert g_k\rVert_\infty.
    \label{eq:app-attention-bound}
\end{equation}
This bounds the dot-product contribution to the attention logits before softmax; any separate additive attention bias must be considered independently.

For fixed gains, the bound is independent of the raw query and key norms. Exact invariance to positive rescaling holds for nonzero inputs when $\varepsilon=0$. With $\varepsilon>0$, it holds approximately when $\varepsilon$ is negligible relative to the mean squared magnitude of both the original and rescaled inputs. Learned gains can nevertheless grow, and attention may remain concentrated. These bounds motivate examining feature scales and attention entropy, rather than establishing that query--key RMSNorm prevents instability or entropy collapse \citep{henry2020qknorm,zhai2023entropy}.

\subsection{Observation Evidence and Contextual Prediction}
\label{app:observation-analysis}

Hold the codebook fixed, and let $p_\star$ denote the joint distribution induced by the data and the forward corruption process. For token $v$ at position $i$, Equation~\eqref{eq:forward} gives
\begin{align}
    \log p_\star(z_{\gamma,i}\mid y_i=v,\gamma)
    &=
    -\frac{\lVert z_{\gamma,i}-\alpha_\gamma c_v\rVert_2^2}
    {2\sigma_\gamma^2}
    +\kappa_\gamma \nonumber\\
    &=
    \frac{\alpha_\gamma}{\sigma_\gamma^2}
    z_{\gamma,i}^{\top}c_v
    -\frac{\alpha_\gamma^2}{2\sigma_\gamma^2}
    \lVert c_v\rVert_2^2
    +\kappa(z_{\gamma,i},\gamma),
    \label{eq:observation-likelihood}
\end{align}
where $\kappa_\gamma$ and $\kappa(z_{\gamma,i},\gamma)$ are independent of token identity. Since $\lVert c_v\rVert_2^2=s^2d$, the norm-dependent term is also constant across tokens. The remaining token-dependent term is the baseline likelihood shortcut used in Plaid and LangFlow \citep{gulrajani2023plaid,chen2026langflow}.

Let $z_{\gamma,-i}$ denote the noisy sequence with position $i$ removed. Under independent position-wise Gaussian corruption, $z_{\gamma,i}$ is conditionally independent of $z_{\gamma,-i}$ given $y_i$ and $\gamma$. For tokens $v$ and $u$ with positive conditional probability given $z_{\gamma,-i}$ and $\gamma$, Bayes' rule yields
\begin{align}
    \log
    \frac{p_\star(y_i=v\mid z_\gamma,\gamma)}
         {p_\star(y_i=u\mid z_\gamma,\gamma)}
    &=
    \frac{\alpha_\gamma}{\sigma_\gamma^2}
    z_{\gamma,i}^{\top}(c_v-c_u) \nonumber\\
    &\quad+
    \log
    \frac{p_\star(y_i=v\mid z_{\gamma,-i},\gamma)}
         {p_\star(y_i=u\mid z_{\gamma,-i},\gamma)}.
    \label{eq:posterior-odds}
\end{align}
The first term is local observation evidence. The second incorporates the other noisy positions and the token prior, so no uniform-prior assumption is required. With unequal embedding norms, the corresponding difference of norm-dependent terms from Equation~\eqref{eq:observation-likelihood} must also be retained.

The learned vocabulary head receives the entire noisy sequence, including position $i$, and therefore need not represent a context-only prior. At fixed $C$, an unrestricted head also admits an equivalent forward representation with an ungated shortcut:
\begin{equation}
    \ell_{\mathrm{raw}}'
    =
    \ell_{\mathrm{raw}}
    +(a_\psi(\gamma)-1)c_{\mathrm{skip}}(\gamma)z_\gamma C^\top.
    \label{eq:app-shortcut-reparameterization}
\end{equation}
Adding $c_{\mathrm{skip}}(\gamma)z_\gamma C^\top$ to $\ell_{\mathrm{raw}}'$ recovers the gated logits. Thus, the decomposition into contextual logits and shortcut evidence is not uniquely identifiable from the combined logits alone. This forward equivalence does not imply equivalent gradients or optimization dynamics. In a finite architecture, the explicit gate may also affect approximation capacity.

\subsection{The Gradient Path Removed by Detachment}
\label{app:gradient-analysis}

Consider a single position after shortcut warmup. Write $z=z_{\gamma,i}$ and let $\beta=a_\psi(\gamma)c_{\mathrm{skip}}(\gamma)$ be a scalar. The logits, represented as a column vector, are
\[
    \ell=r+\beta Cz,
    \qquad r\in\mathbb R^{|\V|}.
\]
For the unweighted token cross-entropy loss $\mathcal L_{\mathrm{tok}}=-\log p_\theta(y)$, where $p_\theta=\operatorname{softmax}(\ell)$,
\[
    \delta
    =
    \frac{\partial\mathcal L_{\mathrm{tok}}}{\partial\ell}
    =
    p_\theta-\operatorname{onehot}(y).
\]
The direct gradient contribution through the explicit $C$ factor is therefore $\beta\delta z^\top$.

Let $g_E$ and $g_E^{\mathrm{sg}}$ denote the gradients with respect to $\operatorname{vec}(E)$ returned by backpropagation through the original and detached computational graphs, respectively. Define
\[
    J_C
    =
    \frac{\partial\operatorname{vec}(C)}
         {\partial\operatorname{vec}(E)}.
\]
Then
\begin{equation}
    g_E-g_E^{\mathrm{sg}}
    =
    J_C^\top
    \operatorname{vec}\!\left(\beta\delta z^\top\right).
    \label{eq:shortcut-gradient-difference}
\end{equation}
Both gradients are evaluated at the same parameter values, with identical forward computations; the only difference is stop-gradient on the shortcut's explicit $C$ factor. Paths through $r$, through $z$ when the noisy state depends on the codebook, and through any other retained computation remain in both graphs. Batch and sequence averaging introduce their usual scaling factors, and the warmup coefficient can be absorbed into $\beta$. Fixed token weights multiply the corresponding gradient contributions.

Equation~\eqref{eq:shortcut-gradient-difference} follows by differentiating the matrix product and applying the chain rule through codebook normalization. It identifies the removed contribution without determining whether that contribution helps or hinders optimization. In particular, a large $c_{\mathrm{skip}}$ does not by itself imply a large gradient, because $\delta$ depends on the prediction. Relative gradient norms and alignment with the retained gradient provide more direct evidence for an optimization-based explanation.

\subsection{Posterior Means, Geometry, and Self-Conditioning}
\label{app:feedback-analysis}

For any distribution $p$ over tokens, let $Y\sim p$, $X=c_Y$, $\mu(p)=\E_p[X]$, and
\[
    \mathcal R_p(a)=\E_p\lVert X-a\rVert_2^2.
\]
Expanding the square and using $\E_p[X-\mu(p)]=0$ gives
\begin{equation}
    \mathcal R_p(a)
    =
    \mathcal R_p(\mu(p))
    +\lVert a-\mu(p)\rVert_2^2.
    \label{eq:conditional-mean-risk}
\end{equation}
For a scalar weight $w=w(p)$, substituting $a=w\mu(p)$ yields
\begin{equation}
    \mathcal R_p(w\mu(p))
    -
    \mathcal R_p(\mu(p))
    =
    (1-w)^2\lVert\mu(p)\rVert_2^2
    \geq 0.
    \label{eq:app-shrinkage-risk}
\end{equation}
The equal-norm codebook further implies
\begin{equation}
    \operatorname{tr}\operatorname{Cov}_p(X)
    =
    \mathcal R_p(\mu(p))
    =
    s^2d-\lVert\mu(p)\rVert_2^2.
    \label{eq:embedding-variance}
\end{equation}
These identities hold under any $p$, whether or not it is calibrated. They describe conditional data risk when $p$ equals the true token posterior.

Entropy alone does not determine the embedding variance. For $d\geq2$, let $b_1,b_2$ be orthogonal unit basis vectors and set $R=s\sqrt d$. Consider equal-norm embeddings $c_1=Rb_1$, $c_2=Rb_2$, and $c_3=-Rb_1$. Two distributions supported on these tokens, with probabilities $(1/2,1/2,0)$ and $(1/2,0,1/2)$, both have entropy $\log 2$, but their embedding variances are $R^2/2$ and $R^2$, respectively. Entropy-based gating therefore provides a heuristic for auxiliary-input weighting, rather than an exact measure of uncertainty in embedding space.

In \ours{}, confidence weighting changes only the training bootstrap input. Both the inference solver and its self-conditioning input retain the unweighted mean. The risk identities above concern a single estimate under a fixed distribution; their implications for iterative generation must be assessed separately. This distinction motivates evaluating auxiliary conditioning and state estimation as separate aspects of the model.

\section{Implementation Configuration}
\label{app:config}

Table~\ref{tab:implementation-config} summarizes the shared training configuration and evaluation protocols for the cumulative architectural ablation. The 250k-update budget is the configured training target; each reported measurement is associated with its recorded checkpoint step. The shared-schedule ablation and the separate extended-training checkpoints are reported separately.

\begin{table}[tbp]
\centering
\small
\caption{Shared configuration and evaluation protocols for the cumulative architectural ablation. Sample counts and actual checkpoint steps accompany the reported measurements.}
\label{tab:implementation-config}
\begin{tabularx}{\linewidth}{@{}l>{\raggedright\arraybackslash}X@{}}
\toprule
Item & Value \\
\midrule
Data & OpenWebText; document BOS/EOS markers; packed random crops \\
Tokenizer & Qwen3 tokenizer \\
Configured vocabulary size & 151,936 \\
Generator weights & No pretrained language-model weights \\
Sequence length & 1,024 tokens \\
Backbone & Width 768; 12 blocks; FFN width 3,072; 12 MHA heads; head dimension 64 \\
Position encoding & RoPE with base 10,000 \\
Dropout & 0.1 \\
Diffusion state & Learned token embeddings with row norm $0.1\sqrt d$ \\
Self-conditioning frequency & 25\% of training examples \\
Configured training budget & 250k optimizer updates \\
Global batch size & 512 sequences \\
Training precision / seed & bfloat16 / 42 \\
Optimizer & AdamW; weight decay 0.01; gradient clipping threshold 1.0 \\
Learning-rate schedule & 5k-update warmup to $3\times10^{-4}$; cosine decay to $3\times10^{-5}$ \\
Evaluation weights & EMA; decay 0.999 \\
Training-time evaluation & Every 10k updates; 1,024 samples; length 1,024; 128 NFE \\
Additional checkpoint evaluation & 1,024 samples; length 1,024; 128 or 1,024 denoising steps \\
GenPPL evaluator & Frozen GPT-2 Large; corpus-level token aggregation \\
\bottomrule
\end{tabularx}
\end{table}

The denoising pipeline is described in Section~\ref{sec:overview}, with its components detailed in Sections~\ref{sec:architecture-backbone}--\ref{sec:architecture-self-conditioning}. For variants using RMSNorm, all normalization reductions are performed in FP32 with $\varepsilon=10^{-6}$. In R3--R7, query--key RMSNorm is applied after projection and head reshaping, before RoPE. Each layer has separate query and key gain vectors of dimension 64, shared across attention heads.

Entropy-based confidence weighting is introduced only in R7. It acts on the bootstrap input for self-conditioned training examples. During inference, both self-conditioning and the solver update use the unweighted estimated posterior mean.

\subsection{Separate Extended-Training Runs}
\label{app:extended-training}

The extended LangFlow + \ours{} result is evaluated at 660k updates after continued training of the full architecture under a separate 1M-update plan. The configured decay is indexed from step 270k, with 2,500 warmup updates, peak learning rate $3\times10^{-4}$, and cosine decay toward $3\times10^{-5}$ at step 1M. Training resumes from the 450k checkpoint, while retaining the decay schedule indexed from 270k. Global batch size is 512, weight decay is 0.01, and EMA decay is 0.999. These settings are separate from the 250k ablation schedule.

The extended Plaid + \ours{} result is evaluated at 520k updates. Its initial training stage uses a 1M-update target with 2,500 warmup updates followed by constant group learning rates: $3\times10^{-4}$ for the backbone/projections and $10^{-2}$ for the embedding and schedule groups. The continuation resumes at 250k, uses 5,000 warmup updates, and then applies cosine decay to 0.1 times each group's peak rate by step 1M. AdamW uses betas $(0.9,0.999)$, zero weight decay except 0.1 on the noise-schedule bounds, and gradient clipping at 1.0. Both stages use batch size 512, bfloat16, seed 42, and EMA decay 0.999. The OWT pipeline and tokenizer match those described in Section~\ref{sec:settings}.

All extended-training results in Table~\ref{tab:main-results} use plain, unweighted inference self-conditioning. The release metadata records the source run, actual checkpoint step, planned target, and sampling mode separately. The selected results do not imply that the 1M-update targets have been reached.

\section{Ablation Configurations}
\label{app:runs}

Table~\ref{tab:run-feature-map} summarizes the seven cumulative configurations used in Table~\ref{tab:ablation} and the training trajectories in Figure~\ref{fig:tb-nfe128-curves}.

\begin{table}[tbp]
\centering
\small
\caption{Cumulative architectural changes. Each row retains the changes introduced in preceding rows.}
\label{tab:run-feature-map}
\begin{tabularx}{\linewidth}{@{}l>{\raggedright\arraybackslash}X@{}}
\toprule
Configuration & Change relative to the preceding row \\
\midrule
R1 & Baseline \\
R2 & RMSNorm \\
R3 & Query--key RMSNorm \\
R4 & 256-dimensional noise conditioning \\
R5 & Likelihood-shortcut detachment \\
R6 & Learned likelihood-shortcut calibration \\
R7 & Confidence-gated training self-conditioning \\
\bottomrule
\end{tabularx}
\end{table}

R7 is the full \ours{} configuration illustrated in Figure~\ref{fig:final-architecture}, with an explicit 250k-update training budget. The 1M-update configuration is treated as a separate experiment. In the cumulative ladder, R4 widens noise conditioning to 256 dimensions, and R5 adds shortcut detachment at that width.

The 50k-step column of Table~\ref{tab:ablation} uses the GenPPL scalar recorded at optimizer step 50,000 for each configuration. These values are neither minima nor averages over earlier checkpoints. The 250k comparison in Table~\ref{tab:main-results} uses R1 and R7 under the same 250k schedule.

\section{Evaluation and Reporting Details}
\label{app:reporting}

Matched ablations compare the same checkpoint steps, EMA conventions, generation seeds, sampling temperatures, NFE, and sample counts. The prescribed generation length is 1,024 tokens. Both the logged trajectories and the additional checkpoint evaluations reported here use 1,024 samples.

\paragraph{Metric definitions.}
GenPPL is the exponential of the mean negative log-likelihood aggregated over scored tokens, as defined in Equation~\eqref{eq:genppl}. Unigram entropy is averaged across individual samples. For $M$ evaluated samples with empirical token-frequency distributions $\widehat q_m$, this is
\begin{equation}
    H_{\mathrm{sample}}
    =
    \frac{1}{M}
    \sum_{m=1}^{M}
    \left[
        -\sum_v
        \widehat q_m(v)\log\widehat q_m(v)
    \right].
    \label{eq:app-reporting-sample-entropy}
\end{equation}
Each $\widehat q_m$ is normalized within its sample; samples are not pooled into a single token-frequency distribution. We refer to this metric as mean per-sample unigram entropy.

\paragraph{Sample accounting.}
The reported checkpoint metrics correspond to files containing 1,024 generated records. Texts shorter than two evaluator tokens contribute no next-token predictions to GenPPL; entropy is computed on their non-padding token sequences. The inline scorer's empty-clean-text fallback is described in Section~\ref{sec:settings}.

\paragraph{Logged-scalar extraction.}
The training curves in Figure~\ref{fig:tb-nfe128-curves} and the 50k-step results in Table~\ref{tab:ablation} are extracted from the training logs for R1--R7. The recorded hyperparameters specify EMA weights, 1,024 samples of length 1,024, 128 NFE, temperature 1, generation seed 0, and the same frozen GPT-2 Large scorer. The generation seed is distinct from the training seed of 42. The tokenizer and special-token policy are specified in Section~\ref{sec:settings}; unigram entropy uses natural logarithms.

The exporter preserves exact optimizer-step indices and raw scalar precision in CSV files, records source-event paths and snapshot checksums, and rounds only the displayed values. Missing TensorBoard measurements are not imputed from other evaluations. These logs are reported with their actual checkpoint steps and evaluation settings.

\paragraph{Sampler provenance.}
R7 applies confidence weighting to the training bootstrap input while retaining the baseline unweighted self-conditioning rule during sampling. A model implementation that also weights inference-time feedback defines a different sampling setting and falls outside the designated R7 comparison. Model class, code revision, sampler implementation, and feedback rule are therefore needed to identify an evaluation unambiguously.

\section{Limitations and Further Controls}
\label{app:controls}

The cumulative ladder measures successive changes in a particular order. It does not by itself establish the contribution of each component in the full architecture or resolve interactions among components. Further controls would include removing individual components from the full model, crossing query--key normalization with conditioning width 128/256, and crossing shortcut detachment with calibration at a fixed conditioning width. A backbone expansion with a comparable parameter increase, together with a separate compute-matched comparison, would further clarify whether gains arise from the placement of conditioning capacity.

The roles of the two gates also warrant separate tests. Comparing a fixed unit multiplier, a learned constant, and a learned noise-dependent multiplier would isolate the value of noise-dependent shortcut calibration. Comparing confidence gating with no gating, constant attenuation, and gates shuffled within noise bins would distinguish token-specific confidence information from generic attenuation. These controls are not included among the reported results.

Such comparisons would require matched data exposure and optimization schedules, comparable tuning budgets, and explicit accounting of parameter counts and FLOPs. Independent baseline and modified-model runs across a shared seed set would assess training variability; intervals obtained by resampling generated texts do not capture that source of uncertainty.

Potential failure modes include suppressing useful uncertain self-conditioning inputs or removing shortcut gradients that aid optimization. Lower feature variance alone does not establish better generation. These possibilities limit the generality of conclusions drawn from the present configurations and motivate evaluating the interventions across broader settings.

\clearpage
\section{Qualitative Generations}
\label{app:qualitative}

We present some unconditional generations covering every checkpoint and sampling budget shown in Figure~\ref{fig:owt-teaser}. Panel (a) contributes eight baseline/modified evaluations from LangFlow and RePlaID within 250k updates; the RePlaID pairs differ in checkpoint progress, and its modified 250k checkpoint belongs to a 1M-update plan. Panel (b) contributes four extended-training evaluations from LangFlow at 660k and RePlaID at 520k, both from 1M-update plans. Table~\ref{tab:main-results} reports the corresponding checkpoints and evaluation results.

\paragraph{Example selection.}
From each of the twelve evaluation files, we display the fixed sample IDs. The evaluation settings follow the twelve bars in Figure~\ref{fig:owt-teaser}; individual samples are not ranked by perplexity, fluency, or perceived quality. The GenPPL shown on each card describes its full 1,024-sample evaluation, not the displayed text alone; the paired entropy is reported in Table~\ref{tab:main-results}. The examples follow the figure's panel and bar order.

\paragraph{Sampling settings.}
All examples use EMA weights, length 1,024 model tokens, temperature 1, final argmax decoding, and unweighted inference self-conditioning. LangFlow uses Euler--EDM with 128 or 1,024 total denoiser calls. RePlaID uses ancestral DDPM with 128 or 1,024 reverse transitions plus one final call, for totals of 129 or 1,025; the card headers denote this extra call by ``(+1).'' The generation seed equals the sample ID. Equal recorded seeds do not imply semantic prompts or establish identical initial noise across formulations.

\paragraph{Reading the examples.}
The complete decoded text is reproduced without correction or truncation, including special tokens and document boundaries. Names, quotations, citation-like strings, repetitions, and factual errors are part of the generated output. The nonprinting code points U+0097/U+009D, the source replacement character U+FFFD, and an unsupported ribbon emoji are displayed by code point (e.g., \texttt{[U+1F380]}). U+200B retains zero-width spacing and U+200E retains left-to-right directionality. Other changes are limited to typographic escaping and layout. These fixed-ID examples illustrate the cited evaluations but do not establish typical quality, factual accuracy, or systematic improvement with NFE.

\clearpage
\refstepcounter{subsection}
\label{sample:langflow-baseline-250000-1-nfe128}
\addcontentsline{toc}{subsection}{\protect\numberline{\thesubsection}LangFlow baseline, 250k: sample 1, 128 NFE}
\begin{samplebox}{\thesubsection\enspace LangFlow baseline}{Sample 1\enspace\textbar\enspace 128 NFE\enspace\textbar\enspace GenPPL 64.26}
{\rmfamily\fontsize{7.5}{9}\selectfont\color{sampleink}Unedited model output\enspace\textbar\enspace 250k checkpoint\enspace\textbar\enspace EMA\enspace\textbar\enspace Seed 1\enspace\textbar\enspace Temperature 1\enspace\textbar\enspace 1,024 model tokens\par}\vspace{5pt}
LMANE FLADA: They have the ability to bomb [the terrorism agent] nearly one way. And, you know, at every process, law forces can bomb almost every time. They just have windows around every process\textemdash{}they can\textquoteright{}t even go anywhere. And they also have the ability of law enforcement to destroy 100 percent of the market, jobs, and everything\textemdash{}some of that has window doors, walls, all go in. All of them.

AMY GOODMAN: On Public Action Network, labor worker Jaime Carr quoted this claim. You know\textemdash{}what they are saying was, \textquotedblleft{}U.S. Border Patrol is illegally.\textquotedblright{} So they\textquoteright{}re basically right, they have done a lead in stopping this immigration problem, even if undocumented workers are dangerous at work.

CLEARTOTT: We\textquoteright{}re also watching this today\textemdash{}there\textquoteright{}s Kate Cole, one of our guest directors of the CISE Institute. Here\textemdash{}we\textquoteright{}ve talked with it\textemdash{}you have a look\textemdash{}in the transcript of one very many immigrants who\textquoteright{}ve come to move out of their country. And all, but are you scared [to put] papers on your head?

LMANE FLADA: You can read it on a big screen, buy a computer; you can write it on a message; you can put it anywhere and you can read it somewhere else. I mean, what you see is\textemdash{}you know, there\textquoteright{}s a very small community of illegal immigrants. And here, what we\textquoteright{}re talking about here, you know, Mexican think of a disastrous border problem is that there\textquoteright{}s about 140 million Americans at risk if our border isn\textquoteright{}t financially harmed. I don\textquoteright{}t care about stories like that. I don\textquoteright{}t know how this works\textemdash{}I mean, we deal with a policy\textemdash{}which in fact we believe is the most important and important benchmark in this century between border safety and the protection of the rights of American workers. I think Kate draws thoughtful feeling of complaints about how being valued is often found here in the United States.

LMANE FLADA: We didn\textquoteright{}t bring jobs here, we got that well. Because it was a little belly\textemdash{}we made a 12 percent wage at the hotel where we operated. And that was good. And again, it\textquoteright{}s not that bad at the Border Patrol, because we did\textemdash{}they had triple the same rate\textemdash{}but people made it profitable that when we were paid about a percentage behind, we didn\textquoteright{}t even feel\textemdash{}they kept the same wage against if somewhere else was. Families still pay high as well. And you know, with the good handbacks we really didn\textquoteright{}t feel around just hammered. And the businesses still pay less. And, apparently, you know it\textquoteright{}s customers good, but the Mexicans in the past sometimes forget there\textquoteright{}s no more for them. And the industry was paying quite the less. This kind of a discrepancy. This is accelerating a recession. So we kind of thought about being paid for a book and then getting out on the circuit of the machine. So the economy is climbing, and the workers are getting lower, and the supply are getting higher\textemdash{}and even more price{-}wise. And so, it means half overall, because this country, it\textquoteright{}s growing. So we underestimated how the United States could\textemdash{}they got differently. We look at the landscapes which Mexico is dreaming about as an example of what\textquoteright{}s happening.

AMY GOODMAN: I would like to go ahead and make a lead up in the end of the release of \textquotedblleft{}RACTLY IN HIS LAVE,\textquotedblright{} that the CISE Institute announced yesterday by Kate, watching, and then directed by Rob Romano. Let\textquoteright{}s be set, and we look forward to talking with you, Kate about what President Trump wants and why this is unfolding. Let\textquoteright{}s rephrase, Republican members of Congress wanted to get it wrong yesterday night\textemdash{}and they wrote about it illegal to the Bill of Justice\textemdash{}is that President Trump wants to communicate with American workers in the border. He\textquoteright{}s going to kick off food to China, or Mexicans and just talk at any number of people. But what American workers did say to me and got you aside was \textquotedblleft{}We can make jobs safe and we have open space\ldots{}and we have our workers..\textquotedblright{} Let\textquoteright{}s point out that this president is in what is part of what he believes has been a good gap between China and the United States. He\textquoteright{}s imposing tariffs on Chinese goods. And so, as President{-}elect he wants to distinguish between people and neighbour who is also the source of U.S. goods. So he wants to sell products from Mexico to China, or at least to Mexico, first. And he is just playing this particular kind of employer{-}agory{-}thanplier approach, something that\textquoteright{}s designed to improve the U.S. economy. He wants to ensure the economy is clean, the markets are building up. Essentially, he\textquoteright{}s label China as a knock{-}win border competitor\textemdash{}and he should be able to get our workers at the border\textemdash{}and that\textquoteright{}s what China should be doing. And, you know, he\textquoteright{}s willing to stop business with Chinese automakers, and, at this point, have some 10{-}second duration. All those companies
\end{samplebox}
\clearpage
\refstepcounter{subsection}
\label{sample:langflow-250000-2-nfe128}
\addcontentsline{toc}{subsection}{\protect\numberline{\thesubsection}LangFlow + \ours, 250k: sample 2, 128 NFE}
\begin{samplebox}{\thesubsection\enspace LangFlow + \ours}{Sample 2\enspace\textbar\enspace 128 NFE\enspace\textbar\enspace GenPPL 53.66}
{\rmfamily\fontsize{7.5}{9}\selectfont\color{sampleink}Unedited model output\enspace\textbar\enspace 250k checkpoint\enspace\textbar\enspace EMA\enspace\textbar\enspace Seed 2\enspace\textbar\enspace Temperature 1\enspace\textbar\enspace 1,024 model tokens\par}\vspace{5pt}
 cannot reserve the opportunity to review and disclose what were now written in order to celebrate "the number of persons condemned ... identified as struggling people [1]" Bell claimed.[4] Despite her friendly talk to Governor Davis, Bell had not announced that she would meet with Governor Davis; but before noon May 24 she announced her intention not to sign on to repeated much time,[5] by arbitrarily traveling directly abroad in San Diego. Bell revealed that Sunday May 27 had not yet finished and, by her most private admission, had she decided to testify as Governor Wise were not meeting. Her statements were in reverse, as Bell had been questioned by the pretrial to Governor Davis and had failed to recognize other key motives on her case, including Robert Wise.

Execution [ edit ]

On February 22, 1961, Bobby Wise, arrested on the charge of felony murder, had been killed in a single, seven{-}blown murder that killed 61 people in Hagerfield. Assistant Attorney General Louis H. Bunsdale announced the conviction in a strongly dissent, joined by then U.S. District Judge William Featherstone.

Sentent in court [ edit ]

A list of several activists and court makers from throughout United States and throughout America in other town included Thomas Willick Thomant, who was skeptical for Judge Featherstone, and Marvin Canoglin, the assembled group of attorneys who assisted Bell. In a starting arguments on May 21, 1961, the judge had account of the long history of criminalities, in whom and when Bell had been involved. Dellegible parts of Bell's arrest were also described as a petty situation in which he lived in licker conditions, drank with poor families, and obliged his wife to sail his way home. 'The contreme' was a test in which Bell reserved her to a "sweet, boiled secret", which he said gave the victim "an impression of error \ldots{} [sealcy] the evilness of the present . . . an isolated, nobely rich person who had such a deliberate and involuntary life".[5]

In exchange for his trial with Thombs, Featherstone reduced Bell's earlier four{-}year sentence and suspended a substantial driving sentence for the following sentence: "For the past five to, 40 years, on that day, Bobby Wise had served 27 hours in solitary ... killing a person suffering from a form of poverty that polluted his physical and mental productivity, leaving him ill, and his ability to do his job. In exchange for a life sentence, Bell had served on the military{-}deaster Lackland Naval Station at Rocky Gap, Utah, Canada and US."[5][6]

Release [ edit ]

Robert G. Starr sentenced Bell to death which he expressed in his petition remarks:[4] "I have already anneded my sentence reserved to the punishment that accomplice C. Bell has perverted, . . . [and] has suffered no repentance at standencely..."

Judge Featherstone's dissents for the Eighth District of States' Court of Appeals determined Bell's acquittal not to exist, and stated that "[it] alleged that this defendant ... made gross remorse in her death and plotted ansacrific destruction of her own property ... with reckless injury to adequately determine the employative effects of her life.\textquotedblright{}[3] The judge agreed with Starr and agreed to leave the sentence in the midst of a next spring recess.

Racial bias [ edit ]

In the other three high{-}profile trials, Bell was charged with murder, grisly shooting death of an 18{-}year{-}old black woman in Lawrence, in other, racially motivated homicides. Bell's first involuntary trial of her admissions was conducted on December 30, 1971. In that trial in Lawrence, Bell died of a inured hemorrhage that burst five minutes after the whose of left arteries in her left arm and left her in full suffocation\textemdash{}then being evaluated and delivered to a hospitalized local.[6]

Judge Featherstone's decision in a unique and divided opinion indicates that the brutality the trial led to. There was police at the scene of the shooting denying warrants and threatening to arrest Grando Mariszewski, a 16{-}year{-}old male, and county law enforcement. But they both suspected Bell of endangering her at the same time. Mariszewski escaped and fled to the streets of both Austin, Berkeley and Point Resille, 14FK) south of downtown. Bell was accidentally shot in the left forehead in the scene.[4]

Release [ edit ]

In the age of the Robert Wise tragedy, the two sides manifestly and adamantly rejected the fact that this incident occurred of obvious consequences. As the majority of Americans\textemdash{}few more than one in all American adults were not\textemdash{}included the treatment of the nearly 100,000 African Americans who had been killed, they conceded that the two factors were an equal consequence. Their fraught right{-}wairings and escalating tensions with the white community undermined the underlying context of violence
\end{samplebox}
\clearpage
\refstepcounter{subsection}
\label{sample:langflow-baseline-180000-1-nfe1024}
\addcontentsline{toc}{subsection}{\protect\numberline{\thesubsection}LangFlow baseline, 180k: sample 1, 1024 NFE}
\begin{samplebox}{\thesubsection\enspace LangFlow baseline}{Sample 1\enspace\textbar\enspace 1,024 NFE\enspace\textbar\enspace GenPPL 41.98}
{\rmfamily\fontsize{7.5}{9}\selectfont\color{sampleink}Unedited model output\enspace\textbar\enspace 180k checkpoint\enspace\textbar\enspace EMA\enspace\textbar\enspace Seed 1\enspace\textbar\enspace Temperature 1\enspace\textbar\enspace 1,024 model tokens\par}\vspace{5pt}
\textemdash{}CATANDMA: They are out of all raids [for that agency] in one way. And, you know, at every event, ICE forces these helicopters almost every day. They patrol these facilities almost every week\textemdash{}they can\textquoteright{}t even go anywhere. And they also have the numbers of law enforcement personnel killed in raids, which actually represents the number of patrols, so situation\textemdash{}some of that is routine enforcement, which will all be raids. Most of them\textemdash{}what we call the Homeland Action Network, law enforcement is also classified as raids. And just\textemdash{}what we are rejecting is not the amount of time law enforcement is attacked. When they are faced up, they are considered a national threat, a national threat, if law enforcement procedures are directed at airports.

GROWD: We\textquoteright{}re also watching this today\textemdash{}there\textquoteright{}s Dale Miller and one of our guest studios, The National Access Network. Here, I\textquoteright{}m with Jacob Miller, a lecturer at the University of South Carolina, who I hope speaks to the people of this country. And here, and welcome you.

GROWD: Here, Jacob Miller.

CRUOW: Thank you for sharing it. It\textquoteright{}s a great eye welcome to Jacob Miller here.

GROWD: Thank you for going it. And I want to tell you today. I mean, what you see is\textemdash{}you know, I\textquoteright{}m a very big fan of illegal immigrants. And here, what we\textquoteright{}re talking about here, you know, the think of a disastrous border security is that there are about 140 million immigrants in this country that we can\textquoteright{}t financially commit. I don\textquoteright{}t care about numbers like that. I don\textquoteright{}t know what those impediments are, we deal with a policy\textemdash{}which in fact we believe is the most important and important benchmark in this century between border safety and the safety of the American people. Don\textquoteright{}t you think the backdrop for that of Donald Trump has been and is often found right in the border policy debate?

CATANDMA: I don\textquoteright{}t know\textemdash{}you know, we got that well. Because it was a little compliment\textemdash{}we saw a 12 percent decrease in the construction of our Mexico border after that was concluded. And again, it\textquoteright{}s not that bad at the border{-}security debate because they knew they had built the wall. But I made it clear that when we were talking about a chart alone, we didn\textquoteright{}t quite really lean toward the new campaign wall. I mean we did build the border{-}wall as well. And then look, building the campaign wall and we actually didn\textquoteright{}t talk around just passed. And as business cycles come now. And for the first 30 years in the history of the economy in the Obama administration, there has been more risk.. And frankly, these are draining the border. This is hurting our economy. This is hurting our economy. So when I\textquoteright{}m talking about Trump, I\textquoteright{}m saying that he\textquoteright{}s actually been the genius of the imagination. But the economy is going down, the jobs are going down, and the imports are just going up, the trade deficit is just unstoppable, it is coming in, it\textquoteright{}s down, it\textquoteright{}s up. And that\textquoteright{}s why the United States is going down anymore. And look at the record, Mexico is going down as well. And that\textquoteright{}s why yes, again, I would like to go ahead and make a lead point in this time\textemdash{}

GROWD: Let me just run back to this about Donald Trump.

CRUOW: Come on, going on. But let\textquoteright{}s leave Jacob Miller. Let\textquoteright{}s say he\textquoteright{}s given the right back to the\textemdash{}Well, let\textquoteright{}s assume Donald Trump is and demand the right thing. Let\textquoteright{}s give the right back to the Republicans. But let\textquoteright{}s say he said\textemdash{}and I wrote about it illegal to the Bill of Justice\textemdash{}is that he really needs jobs to put our jobs on the border. He\textquoteright{}s going to bring back jobs to Mexico, or Mexico and just talk at a number of that. And, you know he, told me and got you together, \textquotedblleft{}We can make jobs safe and build the border wall while we have our jobs.\textquotedblright{} And let\textquoteright{}s point on that we have good markets. Throughout his Administration, Donald Trump has made a great difference between Mexico and the United States. He\textquoteright{}s imposing tariffs on imported lumber. And he pushed 15 percent tariffs on lumber in the same way that he expanded the market of U.S. lumber. So he wants to sell that from Mexico to Mexico, but then actually to Mexico, Mexico. And bring that out. And other kind of worker{-}agress seer policy, something that\textquoteright{}s going to help the U.S. economy. He wants to ensure the economy is steel, the markets are building up. Well, he\textquoteright{}s building a 15{-}mile border wall And though we want jobs to put our jobs at the border, that\textquoteright{}s going to help Mexican manufacturers. And, you know, he\textquoteright{}s got to leave NAFTA with a good market, and, at this point, build a 10{-}mile border wall along the border
\end{samplebox}
\clearpage
\refstepcounter{subsection}
\label{sample:langflow-180000-2-nfe1024}
\addcontentsline{toc}{subsection}{\protect\numberline{\thesubsection}LangFlow + \ours, 180k: sample 2, 1024 NFE}
\begin{samplebox}{\thesubsection\enspace LangFlow + \ours}{Sample 2\enspace\textbar\enspace 1,024 NFE\enspace\textbar\enspace GenPPL 24.04}
{\rmfamily\fontsize{7.5}{9}\selectfont\color{sampleink}Unedited model output\enspace\textbar\enspace 180k checkpoint\enspace\textbar\enspace EMA\enspace\textbar\enspace Seed 2\enspace\textbar\enspace Temperature 1\enspace\textbar\enspace 1,024 model tokens\par}\vspace{5pt}
\textquoteright{}ll reserve the resources to review and clarify what were first published on Friday.

ADVERTISEMENT

Sessions\textquoteright{}s statement came as Trump faced a growing day hunt amid reports that he wiretled multiple Russian nationals to talk to President Donald Donald John TrumpREAD: Trump heads to meet with Senate GOP to fight before 2:34 Cohen Report: Trump vows to press on to fight hard despite and's 'blue' migrant julls in interview MORE. Politico revealed that Trump had called with the vice president, Mike Pence, to spend more on sharing and had originally decided to call the White House.

The meeting came two days after Trump was grilled after he announced a proposal by the House Oversight and Means Committee that would compel the special counsel to focus on any matter, including Mueller\textquoteright{}s investigation into alleged interference in the November 2 election.

Yates refused to comment on the issue, directly from the president\textquoteright{}s office.

Trump had previously called for a congressional probe of possible Trump meddling that provoked widespread outrage from conservatives alike.Trump and Schumer told lawmakers last week to allow Comey to look into the matter in a special response, approved by Congress on Thursday.\textless{}\textbar{}im\_end\textbar{}\textgreater{}\textless{}\textbar{}endoftext\textbar{}\textgreater{}Cardope Benedict

Cardope Benedict, Italian: [c\texttt{[U+031A]}\allowbreak{}s\texttt{[U+0281]}\allowbreak{}y[b]; 17 January 1933)[1] is an Italian religious leader who is one of Italy's foremost conservative and liberal monasteries, having delivered hundreds of documents to many individuals and societies throughout his lifetime. Since he was written in 1917, he has been an enduring witness to many individuals and organisations, including politicians, socialists, and others. Defruised by Benedict after his death, he acted as a man in the traditional Roman Church, having covered years of suffering for many families, and in his zeal to sail his way through Jesus' encyclical.

Early life [ edit ]

He was born to a Donona, in Germany, where he both meant Christian (Germania, c. 1922). His early immersion in the Socratici tradition was characterized by Catholicacy and resistance to social justice. He eventually walked into Italy in the 1920s after Benedict refrained marriage to abortion.[1]

Three years later, he remained in the church until his resurrection. He has been a profound witness to many individuals and organisations, including politicians, socialists, and others. In his humble life, he has received a public dislike from several monasteries, having received considerable support from many Catholics in his life, and his desire to continue his ministry.[1] At the age of 8 he preached on the ground{-}alarge of The General Metabmones: Church, Society and Anceise of Christ's Society (ACTI, G. Print/CLG) which he describes in his book The Warrior: Christ's Soughest Adiled Son and Friends Bleed by Nicampaomo Cretti (Supplement, No. 54), 4 January 1932.[1]

Religious [ edit ]

Born at the age of 14, Benedict was conscious of the church's zeal to persist lies into outrageous views.[2] His first lastdaid was secularism. Father Benedict carefully saw the sordid nature of his own public discussion and he was able to convince his adherents to accept his own teachings.[3] Finly three, he was sent to Roman Catholic Church in Sicily.[4] While Pope Benedict passed away, he began to deny the faithful, overturned their studies, and began collaborating with them. The church ran away and by 1956 he took a deep break in the political circles to better correct him.

At the age of eleven, he was sent to a Roman Catholic Church in Sicily.[5] When he aged of seventeen, multiple life warrants relating to the beatings of Em Mellofine, a 7{-}year{-}old girl, and then alleged to have kicked them both and physically molesting them at the same time.[6]

Father Benedict crossed the rules in both directions. When he aged of eight, he was sent to a Catholic Council in Sicily,[7] multiple life warrants relating to the beatings of Eve Henhase, a 16{-}year{-}old girl, and later alleged to have kicked them both and admitted physically molesting them at the same time. Although Benedict lived his life and served as the face of both marriages, he was sent to Saint Mausuci (RO) Roman Catholic Church in Sicily for adoption in the 1997.[8]

Life [ edit ]

In the age of the Holy See, for the first time, he however adamantly rejected the fact that his conduct consisted of biblical consequences. As an reformer, he believed that in the Church, human actions were not merelyinable for matters of nature or or a disgelling of punishments for humans and chastity. However, he conceded that the adult actions were an acceptable punishment for his morally right{-}pionful and radical opponents. He still focused on the pure context of his
\end{samplebox}
\clearpage
\refstepcounter{subsection}
\label{sample:replaid-baseline-40000-0-nfe128}
\addcontentsline{toc}{subsection}{\protect\numberline{\thesubsection}RePlaID baseline, 40k: sample 0, 128 NFE}
\begin{samplebox}{\thesubsection\enspace RePlaID baseline}{Sample 0\enspace\textbar\enspace 128 NFE (+1)\enspace\textbar\enspace GenPPL 52.32}
{\rmfamily\fontsize{7.5}{9}\selectfont\color{sampleink}Unedited model output\enspace\textbar\enspace 40k checkpoint\enspace\textbar\enspace EMA\enspace\textbar\enspace Seed 0\enspace\textbar\enspace Temperature 1\enspace\textbar\enspace 1,024 model tokens\par}\vspace{5pt}
 up with someone unilaterally who might not even want it to end.\textquotedblright{}

Silva has had years getting somewhere excited about what appeals to expect in his Fight Night fighter this postseason. But Kosora doesn't believe that the fight was enough to express his opinions towards McGregor.

\textquotedblleft{}You always see there are people running around with me in this fight, people telling me, \textquoteleft{}They love this fight, it's just sitting on my head, it's hard on my feet,\textquoteright{}\textquotedblright{} he told MMA Bloody Esprit. \textquotedblleft{}Oh, yeah, I definitely want it to happen.\textquotedblright{}

But Kosora said he was appreciative and excited over the fight with Diaz: \textquotedblleft{}There were a lot of possibility, but you can relax and just build a tune and do the same thing that they've done throughout the history of the UFC.\textquotedblright{}

 Kosora still doesn't have a message he made during Ultimate Fighter 17 last October. But while he doesn't blame people here, in polite terms, he still admitted he struggled with just the one fight that complicated the matchup against McGregor.

\textquotedblleft{}It's great to be in MMA because we should be around each other,\textquotedblright{} he said. \textquotedblleft{}Obviously, we have to win on an equal basis, but I think there are some things out there that didn't really have much setup for us. There is a very good chance we need to know what to do.\textquotedblright{}

As for the UFC champion outside of McGregor, Kosora went off the bat regarding this fight. The fighter didn't say they don't yet know it.

Ichewing a microphone off Diaz's right arm, he tapped his grip in McGregor's head and asked Kosora to wait for 30 minutes until UFC lands.

\textquotedblleft{}[People] didn't see it working out,\textquotedblright{} Kosora said. \textquotedblleft{}What they saw about it before UFC and how they felt about it. In combat, we feel like we need more attention than they have to give us in 2015. I don't know if it's ever going to be easy to get more attention, frankly. I don't know if it'll ever get there yet. But we just have to focus on picking up our cards, and we'll focus on competing for this series.\textquotedblright{}\textless{}\textbar{}im\_end\textbar{}\textgreater{}\textless{}\textbar{}endoftext\textbar{}\textgreater{}NORTH LAKE CITY, Va{-}{-} With 8{-}year veteran Robert Ingram wrongly saying he injected him last Friday and died of suspected PTSD from the military, sergeant Van La Buren said his son should go back on trial for criminal mischief. Viewing the video corroborated his comments.

On the morning of Nov. 11, Ingram injected the clinic with alcohol, and delivered the tube to his bedside in a No. 355 classified report issued by the Department of Defense Force Affairs on December 11, 2017, in Fort Napa, Nev. (A+ENews) He received a/send without four gallons of hot water, according to PERF's documents, courtesy of sergeant Van La Buren. (Photo: A\&ENews)

Page. 8{-}Staff Officer Recruiting Director (Mike Ortiz), who fell into a closed lung, told military authorities that he met Van and did not personally rule on the criminal report during a later hearing.

\textquotedblleft{}When Ingram took an escort into custody outside the United States in December 2015, it is clear that international law does not require an administrative tube escort in the first place,\textquotedblright{} Van La Buren said in a statement. \textquotedblleft{}The caller did not know the contents of the tube in any other language, and showed no sign of deception. It is not a documented criminal incident for Title Six, and law enforcement from U.S. Defense and the FBI are concerned.\textquotedblright{}

The sergeant said he transferred the tube to a clinic in New York, where he reportedly underwent an intensive procedure in 2012.

\textquotedblleft{}My mendacity was that we let the case come to us during the Obama administration,\textquotedblright{} La Buren said, noting that the VA was very supportive of the Obama administration\textquoteright{}s conduct and put the tube in custody last year.

And when he asked if it could return to the medical community for now, he initially joked that the VA documents would be used as \textquotedblleft{}repete assurances.\textquotedblright{}

\textquotedblleft{}I\textquoteright{}m very happy with them,\textquotedblright{} Van La Buren said. \textquotedblleft{}I think that all these kids have to wait to learn immediately.\textquotedblright{}

Van La Buren said he would also refuse to bring charges.

\textquotedblleft{}Once we had people of inappropriate information until we spoke, we did not know what\textquoteright{}s at issue for VA,\textquotedblright{} Van La Buren said.

The ex{-}VA captain is beholden to the position of various military leaders, including Lt. Gen. Robert E. Decker.

\textquotedblleft{}We did what we expected and our regret was there weren\textquoteright{}t enough psychopaths on the books,\textquotedblright{} Van La Buren said.

Describing Lt. Morgan as the condition\textquoteright{}s director, he said the facility would reduce the use of its specialties, including psychopaths, by military officials.

\textquotedblleft{}I personally don\textquoteright{}t believe it at the heart of our military component,\textquotedblright{} he said.

Oakland Gov. Barbara Lee traveled from Puerto Rico on Monday to speak
\end{samplebox}
\clearpage
\refstepcounter{subsection}
\label{sample:replaid-150000-0-nfe128}
\addcontentsline{toc}{subsection}{\protect\numberline{\thesubsection}RePlaID + \ours, 150k: sample 0, 128 NFE}
\begin{samplebox}{\thesubsection\enspace RePlaID + \ours}{Sample 0\enspace\textbar\enspace 128 NFE (+1)\enspace\textbar\enspace GenPPL 38.14}
{\rmfamily\fontsize{7.5}{9}\selectfont\color{sampleink}Unedited model output\enspace\textbar\enspace 150k checkpoint\enspace\textbar\enspace EMA\enspace\textbar\enspace Seed 0\enspace\textbar\enspace Temperature 1\enspace\textbar\enspace 1,024 model tokens\par}\vspace{5pt}
 did for \textquotedblleft{}Eergie.\textquotedblright{} However, this amounts to a huge loss/loss episode since this is a story about somebody who couldn\textquoteright{}t ultimately be overutilized for her, even if she never slept with him and was planning on being married/Justified. Oliver Stone exclusively reveals that the producers fell out with Bonnie after they decided to continue working in the role for \textquotedblleft{}Justified.\textquotedblright{} They eventually decided to add Anna Rhett to their base and Bonnie instead hooked up with Renee Kaling (as Bernadette). Walton ultimately moved out to being Alyssa (Trevary Hathaway) and then on the role as Laurie.

\textendash{} Logan got a lot of attention, but she wasn\textquoteright{}t so much conscious of watching and playing a big part as she was as a teenager in \textquotedblleft{}Sherwalk.\textquotedblright{} Jolie eventually decided to stay with Lauren \textquotedblleft{}Liz\textquotedblright{} Fox (as Bernadette). Walt and Jolie were slightly more successful with their audition for \textquotedblleft{}Liz.\textquotedblright{} We\textquoteright{}ll also reveal in this episode that the producers initially reached out with Jolie on the scene with Radar. They were also slightly more successful with their audition for \textquotedblleft{}Eergie.\textquotedblright{} This amounts to a huge loss/loss episode since this is a story about somebody who couldn\textquoteright{}t be overutilized for her, even if she never came with him and was planning on being married/Justified because her primary target was Radar. Jolie also lay the groundwork for this scene on the show from the moment we originally saw it where she played Smoop (Amy Poehler) as Dexter. She dropped her role as Dexter\textquoteright{}s girlfriend and step{-}mate.

\textendash{} October 13, 2011 \textendash{} James Lin on a drive{-}in role in \textquotedblleft{}Try What You Can\textquotedblright{} \textendash{} Lin reportedly agreed to a drive{-}in role that followed the movie. The network signed him late last year for a credit deal. They finally renewed him in 2012. Lin reportedly was returning because that his time on the broadcast television show was more lucrative than ever. Lin then hinted soon after that he might get one last position on the show at the end of the \textquotedblleft{}American River.\textquotedblright{}

\textendash{} Lin reportedly took on a drive{-}in role with Jessica Lange (as Bernadette). She eventually hooked up to Bill Slater as Morgan (Trevary Hathaway) and dealt with his hiatus for another role as Chevy (Amy Poehler).

\textendash{} Lange attracted a lot of attention after she got a few guest voice{-}offs for the show (though not enough to replace The Walking Dead).

\textendash{} Lange attracted a lot of attention after she joined the cast for another role on the show. It was easily her best performance in \textquotedblleft{}Dead Breaker.\textquotedblright{} Lange got a raise of \$200,000 for the Dead Breaker (Portrayor) pre{-}production, but got a raise of slightly less than \$500 for her second role on the show.

\textendash{} October 16, 2011 \textendash{} Jolie goes on a drive{-}in role with Erica (\textquotedblleft{}Donald Glover\textquotedblright{}) as Bernadette \textendash{} Erica, played by Michelle \textquotedblleft{}Walter Bell,\textquotedblright{} isn\textquoteright{}t exactly the same name as AMC in \textquotedblleft{}Justified.\textquotedblright{} Jolie also went on a drive{-}in role at ABC in November (2011). She was hired as Comedy Central\textquoteright{}s talent model/ executive video assistant in the first ever \textquotedblleft{}American River.\textquotedblright{} After losing her real job at the network, she was supposed to be a video assistant for her second role, but didn\textquoteright{}t get the gig. She\textquoteright{}s currently a lifetime talent scout for Comedy Central, which has offices in Los Angeles and Washington. After losing her real job, she decided to join the cast for another role at the network. Jolie attracted a lot of attention in the role after she got a few guest voice{-}offs for the show (though not enough to replace The Walking Dead).

\textendash{} Jolie decided to get close with Pearl since he didn\textquoteright{}t explicitly/homepletely love her during their second scene in \textquotedblleft{}Try What You Can.\textquotedblright{} She also decided to get engaged to Rosa (Josef).

\textendash{} Jolie garnered a lot of attention since Pearl didn\textquoteright{}t implicitly/homepletely dislike her in their second scene. She eventually hooked up being Erica (Walter Bell) and then dating Emily (Lizzie Sedarum). Walt and Jolie were also slightly more successful with their audition for \textquotedblleft{}Justified.\textquotedblright{} We\textquoteright{}ll also reveal in this episode that the producers initially reached out with Yates \textquotedblleft{}Nixon\textquotedblright{} (as Bernadette). They initially decided to bring Anna Rhett to their side and instead reached out to Bill Binder as Erica (Trevary Hathaway). They also decided to stay with Rhett since he wasn\textquoteright{}t under{-}powered. She eventually hooked up with Renee Kaling (as Bernadette) and while she hates getting married, she got the chance to join the cast in this role for \textquotedblleft{}Justified\textquotedblright{} and because of that, they reportedly didn\textquoteright{}t put her in the audition for Bill Gavin. Jolie also went on a drive{-}in role at ABC in November (2011
\end{samplebox}
\clearpage
\refstepcounter{subsection}
\label{sample:replaid-baseline-220000-0-nfe1024}
\addcontentsline{toc}{subsection}{\protect\numberline{\thesubsection}RePlaID baseline, 220k: sample 0, 1024 NFE}
\begin{samplebox}{\thesubsection\enspace RePlaID baseline}{Sample 0\enspace\textbar\enspace 1,024 NFE (+1)\enspace\textbar\enspace GenPPL 23.91}
{\rmfamily\fontsize{7.5}{9}\selectfont\color{sampleink}Unedited model output\enspace\textbar\enspace 220k checkpoint\enspace\textbar\enspace EMA\enspace\textbar\enspace Seed 0\enspace\textbar\enspace Temperature 1\enspace\textbar\enspace 1,024 model tokens\par}\vspace{5pt}
 could without lowering the camera.

(LFB.com) And how many professional cinematographers have spoken with you about the kind of collaborative language they use when their shoots are abroad?

MANA: So far I have reached out to other photographers all over the globe. I have attracted different international cinematographers to my shoot, coming from the most powerful photographic groups in the world. Some international cinematographers have flown in from Canada and the United States over the last few years to pursue opportunities in South America.

(LFB.com) Do you curtail subject selection during your shoot?

MANA: I don\textquoteright{}t have to follow people. I shoot a lot slowly. I have a lot of breakdowns during shoots. I don\textquoteright{}t leave people alone, I don\textquoteright{}t physically leave people alone, so they can take an entire day. But I have to try to communicate and try to keep everyone safe.

(LFB.com) Many skeuaging pictures require shooting transparency. Does your technique differentiate you from viewers?

MANA: Maneuading is really the most challenging thing. Sometimes I hear the least from viewers because I shoot so widely. I don\textquoteright{}t try to blur details \textemdash{} I try to keep certain subjects extraneous. I leave a whole lot of pictures out there that just don\textquoteright{}t give a good picture. It takes me a long time to classify a picture.

(LFB.com) What is the most challenging part of photo taking? How many contactless photos do you use in post? Do you share more than a few?

Paula Hyjic: This is the most difficult thing. I think you have 80 different pictures out there and probably much fewer than Vents About or Dogs Living In. But viewers love transparency and want to use the best details out there, like what they got from learning after university.

(LFB.com) Which other professional photographers have you reached out to on your photo shoots? Do you think you\textquoteright{}ve done the most press shoots in many years?

Paula Hyjic: My goal is to establish new approaches as quickly as possible. A large number of them have come from other cities where I live such as Berlin, Tokyo, and Grenada. Sometimes I shoot in the States.

(LFB.com) With the exception of Jordan Kosar and Andrew Schmidt, who have, for some reason, been quite saturated with talent. There are a few photographers in my division.

(LFB.com) Who is the most representative cinematographer in your team? Has the program begun to form after your meetings with international cinematographers?

MANA: In terms of associating with international photographers, let me say that there are a few ones that I haven\textquoteright{}t met yet, here are the ones that keep track of me. Here are a few ones that I haven\textquoteright{}t heard of enough individuals to know me, and there are others that I haven\textquoteright{}t met. These meetings are basically side trips.

(LFB.com) Which cinematographers have you met previously when it comes to photography?

MANA: Most of the footage and photos are still in there, so that kind of culture I have started to cultivate within months of shooting is important. I have had several conversations with some of my professional photographers, where I ask them questions, explain where they came from, who they are, and sometimes bring them up on my own. These connections have been important to my success going into photo production.

(LFB.com) Which cinematographers do you trust to interpret the photographs you record? How does this depend on what techniques you\textquoteright{}ve learned during filming?

MANA: I want to learn techniques of communication in photography. They have gone a long way in understanding images and techniques that have helped me find new ways to catch the gaze of others.

(LFB.com) Do you collaborate with other focus groups in your close shoots?

Paula Hyjic: Many of these professional photographers want me to travel to meeting venues around the country, such as Boston, New York, amusement parks, Chicago, theatres, galleries, solo pools and the public library. Many of these photographers want me to expand into an artistic center, rather than in a kind of public space.

(LFB.com) What cinematographers have you encountered in your division when it comes to photography?

Paula Hyjic: How many cinematographers do you feel have posed an challenge to other photographers in your division since you first reached out to them?

MANA: I\textquoteright{}ve found 100 different cinematographers in my division. Doug Leelowitz, Edward Kaczynski, James Mann, James Fallon, and Marc McIlroy have all stood out in my team. I\textquoteright{}d say that the best photographer I\textquoteright{}ve encountered is Charlotte Mateo.

(LFB.com) Which international cinematographers have you added to your team?

Paula Hyjic: When working with any of these international photographers, I generally bring out my client\textquoteright{}s work in my entirety. With Frank Deraziano, Christian Head, and David Stange, I take the stories of James Mann, Andrew Schmidt, Doug, Greg Rodriguez, Stephen Bocic, and Jordan Kosar known
\end{samplebox}
\clearpage
\refstepcounter{subsection}
\label{sample:replaid-baseline-220000-1-nfe1024}
\addcontentsline{toc}{subsection}{\protect\numberline{\thesubsection}RePlaID baseline, 220k: sample 1, 1024 NFE}
\begin{samplebox}{\thesubsection\enspace RePlaID baseline}{Sample 1\enspace\textbar\enspace 1,024 NFE (+1)\enspace\textbar\enspace GenPPL 23.91}
{\rmfamily\fontsize{7.5}{9}\selectfont\color{sampleink}Unedited model output\enspace\textbar\enspace 220k checkpoint\enspace\textbar\enspace EMA\enspace\textbar\enspace Seed 1\enspace\textbar\enspace Temperature 1\enspace\textbar\enspace 1,024 model tokens\par}\vspace{5pt}
 is the front tire of a car on fire Sept. 21, 2013. (Photo: Senate Free Press)Buy Photo

WASHINGTON, D.C. {-} A 26{-}year{-}old William Russell was hospitalized after he pushed a stolen van onto a sidewalk Sunday barbed in the 10 block of Lawrence Street near Euclid Avenue. Johnson turned and ripped Russell's head into half, according to a Post police report.

State's Deputy Jesse Schlosser said the 26{-}year{-}old van fell profusely out of the bar then went to the side of the sidewalk, according to the police report.

Police said when Russell fell to the ground, Johnson kicked him in the back and forced him onto the sidewalk. Police said he was in life{-}threatening condition.

Appaloosa County Sheriff's Office Officer Richard Nunn said his car would have been braking when the van fell off the trail.

"I was driving toward the end of the trail when Russell was up there on the ground and pulling up his clothes on his knee," Nunn said.

Schlosser said he got Russell along the trail and drove him under it, then to Tatina Hospital, where he was stabilized in a hospital. He is being held without injuries reported.

Investigators are searching for a Dodge Jeep Cherokee every Sunday from June 30, 2012 to Sept. 21, 2013 to Aug. 11.\textless{}\textbar{}im\_end\textbar{}\textgreater{}\textless{}\textbar{}endoftext\textbar{}\textgreater{}Self{-}modified non{-}digital disks, or the Bionic Killer, are widely used in portable virtual reality devices like computers and smartphones. The virtual reality team maintains its own smart Internet infrastructure, using it to store customer{-}licensed hardware. The hardware's storage and processing systems are configured to instantly transfer older files from an open storage device's SD Card or directly to the Internet.

Note: This story about the Bionic Killer first appeared on the Forbes entertainment website.

WASHINGTON \textemdash{} Orlando police are investigating the exact circumstances of a case in which a west Florida teenager is accused of shooting at three or four school employees at a South Florida high school band club.

A release released June 1 from the Florida Bureau of Investigative Reporting (ICE) says investigators are working to determine how the student was targeted and the school employee is wounded.

Florida District Attorney Triple Green has formally accused the teen of murder. The teen is identified as a man listed in the accounts of four school employees killed in Hurricane Irma earlier this month.

Robert White, a between 1{-}foot{-} and 2{-}foot{-}tall student at Walldale Middle School in Smyrna, South Florida, was killed by gunfire on the school's U.S. campus.

He has been charged with murder, reckless manslaughter and disorderly conduct.

White had worked as a band assistant at the Federal{-}Madland School from 2010 to 2017. White previously led the club's dance room as chief of staff, according to sources close to the school's leader, Federal{-}Madland President PJ York.

No criminal charges have been brought against the school, the Walldale School District, the Smyrna School District or the high school band's president of the district, Ray Beckman.

District Attorney Triple Green said the governor's office is investigating this kind of deadly incident because it's difficult to imagine why someone claiming to possess a gun would make such threats.

In a statement Friday, Federal{-}Madland President and Chancellor Mary Kay Young said the failure to release details is unnecessary and threatens public safety.

"Sometimes a young man violently steals a gun," Young said. "A young man harms his community. Sometimes they don't harm their community."

Several members of the club have been responding to death threats, including 13{-}year{-}old teen Billy Goldman, who appeared on stage on Thursday and demanded that a man killed at the club not be identified by the police.

"Seeing a young man carry a gun is not wrong!" Protestors screamed at an arms rally in Washington, D.C.

"Is seeing a young man carry and hold a gun is okay?" one supporter asked.

District Attorney Green and Mayor Charlie Pines, however, described the release as "ridiculous" and urged further investigation.

The shooting occurred in an upscale field near the club and its main host. The clubs are divided into groups of public school, business{-}school, black and special needs groups.

Florida Police Chief Kevin Evans said his department has been working overtime to assist with the investigation.

Advertisement

Afterward, police formally located the teen, but said they had no reason to disperse him from the scene.

"We saw the shooting, and told him we need to disperse him from the scene," Evans said.

The youth was in black uniformed school officials clothing and armed to conceal their weapons, sources close to the club said.

When officers began a search, they found a young man named D{-}hlik White hiding outside the club, sources said. Rather than hide inside the club, White apparently concealed his band merch, which could increase the likelihood of a mass shooter in the field, sources said.

Police
\end{samplebox}
\clearpage
\refstepcounter{subsection}
\label{sample:replaid-250000-1-nfe1024}
\addcontentsline{toc}{subsection}{\protect\numberline{\thesubsection}RePlaID + \ours, 250k: sample 1, 1024 NFE}
\begin{samplebox}{\thesubsection\enspace RePlaID + \ours}{Sample 1\enspace\textbar\enspace 1,024 NFE (+1)\enspace\textbar\enspace GenPPL 13.90}
{\rmfamily\fontsize{7.5}{9}\selectfont\color{sampleink}Unedited model output\enspace\textbar\enspace 250k checkpoint\enspace\textbar\enspace EMA\enspace\textbar\enspace Seed 1\enspace\textbar\enspace Temperature 1\enspace\textbar\enspace 1,024 model tokens\par}\vspace{5pt}
 was the stepchild of an assist from linebacker Lane Kiffin, who was named the 2014 AT\&T SEC Defensive Player of the Year. Adams recorded just two tackles in his 2014 Big Orange campaign, where he lost a meaningless game to Kansas. He finished third behind Ward (61 tackles at Bryant) and Spurrier (60 in tackles in the month of October).

Adams also made his senior conference debut in the Pinstripe Bowl, playing defensive tackle, left tackle and right guard; three positions that most fans never think to. Adams became a leading talking point with former Alabama coach Mike Herman after a 14{-}point loss at North Dakota State in the third{-}consecutive game of the season, a loss that was largely epitomized by his uncharacteristic play last month.

With Cullen Atkin in serious trouble due to his unfortunate knee injury in last fall, Adams also started his first game for the Golden State Warriors, leaving him in the shadows of many fans' perceptions of his NFL career. Since 2013, Adams has started six of his last seven games for Penn State, beginning in the 25{-}20 loss to Louisville. However, he went on to have a somewhat impressive performance in a Fiesta Bowl game against Georgia on September 30, 2014. He has recorded just one tackle in all three games and three after a 7{-}2 record in last season's clinch conference semifinals versus the University of Arizona, which enabled him to emerge as the program's fifth true freshman.

Adams has also seen a modest amount of success since making his first collegiate snaps for Penn State. The pass{-}coverage and tackling specialist has provided a tremendous completion percentage and a significant early{-}season impact on the Cardinals, forcing the only regular season interception of his career in the Bucksee Stadium victory over Notre Dame. In terms of consistency, Adams continues to be of little benefit as an offensive lineman throughout the Penn State defense's struggles over the past month and half.

Cornerback

Brandon Adams has been the offensive linemen in the Utah{-}Mexico Elite Eight contest against Real Salt Lake in Baton Rouge on Sunday, recording just one tackle in all three games, including nine takeaways in the season{-}opening game against Georgia and the semifinal series against LSU.

Adams has recorded just two tackles for the Bulldogs in all three games, going two for eight in their 62{-}touchdown win over Missouri and in their improbable 26{-}touchdown victory over the Alabama Crimson Tide on Saturday night in Nashville. Adams has recorded two tackles in all three games and nine takeaways in the season{-}opening game against Georgia and the semifinal series against LSU. Adams has also appeared in three starts for the Utah{-}Mexico Wildcats from 2012 to 2014. Adams also filled in as the signal caller for Utah last season, posting an incredible performance for the team Sunday night, in the NACUSA Peach Bowl.

Adams will have significant impact for Penn State in the Pinstripe Bowl, with the defensive lineman filling out a large version of the defense in the Heisman Bowl. Following the NCAA Elite Eight matchup against Real Salt Lake in Baton Rouge on Saturday, Adams has recorded just one tackle in all three games and a sack, throwing his interception in the fourth quarter of the season{-}opening Utah{-}Mexico game against Florida State. Adams will again be the offensive signal caller as the offensive linemen for Florida State, although the only caller ever interviewed for this job was former Florida State defensive lineman Curtis Sims.

As far as his impact on the Penn State offense in the regular season, a bright future for Adams awaits. The 25{-}year{-}old Adams, who went on injured reserve earlier this season while the team dealt with the situation, was named to the Wildcats' varsity squad following Saturday night's emphatic 63{-}touchdown win in the Peach Bowl over Missouri. The Georgia product's quick burst and tremendous skill sets have been an understated contrast to those of his high school team{-}mate Rhone Vokes, hence this is a great opportunity for Adams to add to that comparison as well.

Adams has recorded just one tackle for the Tigers in all three games, including going three for eight in their 33{-}25 Peach Bowl victory over Georgia. LSU athletic director Kelly Kelly announced that she will be signing Adams to transfer to Penn State.

Running Back

Following Saturday night's win over Missouri, Adams has recorded two tackles in all three games, including nine stops and 27 takeaways in the Heisman Bowl to incentivize the return of receiving tackle Jonathan Stewart, who nearly landed on the redshirt.

Adams also anchored the run{-}coverage and tackling defense for the Ducks on Monday night, recording four tackles leading up to an interception offensive lineman Tyler Hightower surrendered in the team's season opener. However, without a similar play to the late Aaron Gordon, the Ducks will likely face the prospect of an explosive dearguard return of their own.

With
\end{samplebox}
\clearpage
\refstepcounter{subsection}
\label{sample:langflow-660000-2-nfe128}
\addcontentsline{toc}{subsection}{\protect\numberline{\thesubsection}LangFlow + \ours, 660k: sample 2, 128 NFE}
\begin{samplebox}{\thesubsection\enspace LangFlow + \ours}{Sample 2\enspace\textbar\enspace 128 NFE\enspace\textbar\enspace GenPPL 44.72}
{\rmfamily\fontsize{7.5}{9}\selectfont\color{sampleink}Unedited model output\enspace\textbar\enspace 660k checkpoint\enspace\textbar\enspace EMA\enspace\textbar\enspace Seed 2\enspace\textbar\enspace Temperature 1\enspace\textbar\enspace 1,024 model tokens\par}\vspace{5pt}
 is out and preparing for a situation in which I consider myself highly, and I\textquoteright{}m hoping that that\textquoteright{}s worked out when I can take the last words off the driver\textquoteright{}s door. Ultimately, I\textquoteright{}ll do all I can until I\textquoteright{}m able to try to get into better PB for a few hours of testing, and will continue to be truly competitive for 2014.\textquotedblright{}\textless{}\textbar{}im\_end\textbar{}\textgreater{}\textless{}\textbar{}endoftext\textbar{}\textgreater{}On Friday, Dec. 1, 2016, Bourlo Sheriff's Office disciplined Kenneth Diamond for refusing to disclose a police recording of the alleged shooting of City Councilman Scott McDonald.

Upon accusing Diamond of "sperring a source of evidence" to the former San Francisco DA earlier this month, Diamond was "very angry."

Then, pre was the video recorded by Officer Kenneth Diamond.

It lowers the substance of Diamond's possible videotape even after the DA's office opened fire on the recording four days ago and filed a Los{-}area lawsuit Thursday saying the recording was not justified and that the officers may have contacted deputy attorney general Maureen Sepan Jr.

The video recorded Dec. 4, 2014, shows a 40{-}year{-}old East Oakland resident arguing with Diamond and with a off{-}duty gun in a Hyundai Volta sedan. The recas indicates that Diamond and Sepan, with the gun in the car and began yelling Sepan at McDonald.

The ensuing incident led to Sepan's usual fisting between the two judges. Confronting Diamond, Diamond told Bob that Sepan had "pointed eyes" and told him that McDonald fired.

Diamond said Bob sportied McDonald several times, then said, "Seth me the fat vomit in my car, Scott McDonald...what are you thinking?"

The slub in the statement, which Diamond described as "very disturbing," prompted a backlash to the Chronicle.

Bob's spokesman said McDonald's lawyers refused to disclose the videotape "for the safety of a client making an uncertain statement about a conviction."

But Berry's DA's office said in a statement: "people in the government's public affairs know that Kenneth has abused his judgment."

Almost three years before the repairup was intervened, Melvin Howard said that "our lawyers have carefully and thoroughly reviewed ... Kenneth's language" and that his apparent violation of Rule 25 "would be unacceptable."

Warren police Chief Bill Bratton said the debate about {-}{-} exactly the coverup would be as the result of a violation is "quite speculative to its part. It's what any slay of actual evidence could potentially have picked."

"There's a lot more unique," Bratton said. "But there's another element that's actually a coverup. Because it almost has to be a murder. You're being shot across the country and you're recording it around the edge, but in court without a search or a warrant, let that happen. That's speculation. It's obviously a coverup."

The DA's office said in the statement of Diamond's lawsuit that it "accepted the court the videotape was unlawful."

They have ruled that the officer's interview was "edited or edited," but had no knowledge of whether the investigation was "serimately filmed in which the officer acted," the office said in the statement.

The incident began as a result of a verbal encounter between former San Francisco DA Ron Sepan and an Alameda Police Department officer on Nov. 9, 2014.

Following several months of checks, court documents and text messages, Diamond filed federal claims against Sepan in the summer of 2016, records show.

Diamond conceded that the suit was "actively provided" and said Sepan was personally liable.

Similar claims were filed through the County Probate Office. But Sepan said he did not intend to legally challenge Diamond's claims. Sepan said Bob was unaware of the lawsuit.

Hours later,'s attorney John Stern announced that the coverup got steroids after the incident.

He moved the full scope of Diamond's lawsuit to the facts of the allegations, saying the recording took place in a minor manner.

That night in December 2014, McDonald told Howard that Sepan stopped him and went to a parking lot, leading McDonald to the tracks and stopped them. He went back for Howard to pursue an attorney and found McDonald screaming in the street. According to the DA's office, Sepan "brested McDonald several times and said, 'Saw the fat vomit in his car.' And he was so scared."

Howard said Sepan paid the city \$365,000 to although with having to assist Bob, who appeared to be wanting to get court order.

The case "made an [untruth]worthy verdict," according to Gov. Jerry Brown. He wrote that she "rined to blend past mistakes with the videos" and monkeyetched up both accusations as if they were apprentice facts."

Deputy Attorney Alisha Deshalka emailed Craig Taylor that she was unaware of the videotape which
\end{samplebox}
\clearpage
\refstepcounter{subsection}
\label{sample:langflow-660000-1-nfe1024}
\addcontentsline{toc}{subsection}{\protect\numberline{\thesubsection}LangFlow + \ours, 660k: sample 1, 1024 NFE}
\begin{samplebox}{\thesubsection\enspace LangFlow + \ours}{Sample 1\enspace\textbar\enspace 1,024 NFE\enspace\textbar\enspace GenPPL 21.72}
{\rmfamily\fontsize{7.5}{9}\selectfont\color{sampleink}Unedited model output\enspace\textbar\enspace 660k checkpoint\enspace\textbar\enspace EMA\enspace\textbar\enspace Seed 1\enspace\textbar\enspace Temperature 1\enspace\textbar\enspace 1,024 model tokens\par}\vspace{5pt}
way. It leads to the Cumberland Estuary without the bank, and water (usually water) passes through Vanville and Mackenzie Lake. It intersects the comparatively steep side of the wall with Oldy Lake to the east of Peterborough, along Annex Avenue. It traces Mackenzie Lake[10] to Prince Rupert, which is the main Canadian river between Vanville and Mississauga.[11] The bridge features high access to the land friendly plants at a width height of 10 metres (14 in in), with a sedimentary foundation with a shallow water surface formed of ice{-}black water and 10 cm (12 in) sand paws, and with a sediment heradow of 10 metres (10 in ft).[12]

History [ edit ]

Route [ edit ]

The Mississauga bridge rises at the base of the Cumberland Estuary. It passes through Peterborough, Wallace Bay and Port Rupert to the west, part of Montreal to the north. From the time the bridge crosses the river, the roadway is seperated from the future. The bridge crosses Toronto and Mississauga and is a section of the Surrey Freeway Line. In 1967, the Montreal River Reservoir line began funneling into the southeast of Mississauga. Water was taken from this route south of Peterborough after a power attraction broke.[10]

The bridge was proposed in 1970 with the support of the Ally Vance Corporation. In 1972, the company decided to forge a bridge based solely on air and hydroelectric to the Huroe Valley.[5] In 1974, the bridge was completed and financed by the federal government. The BIS was tasked with planning an ambitious bridge plan, and the corporation established its Budget and Development Board to help the railway secure contracts. The BIS engaged primarily in grants and expenses associated with economic projects (including funding). As a result the bridges were flooded, with sections of the bridge being sent over to be diverted into the river.

The BIS opened in 1978, but the bridge ran into demolition. In 1978, the bridge was assessed to be flooded and strengthened; in 2002, the BIS placed a moratorium on Nevada Corridor's construction. While a replacement trench has been funded, the bridge was not completed at one time by the end of 2006; by November 2012, the bridge was again considered to have been completed for debt order.[5]

Building [ edit ]

Construction of the Lufferon Bridge and Danarr Drive, the southeast of Mississauga since 1967 broke a volume budget extension was launched in 2013 in Kingston. The funds paid for the bridge were collected from the City of Kingston and the Wewardstein Corporation in conjunction with federal funds. The design of the project was funded along with private funds generated to allow Mississauga to build bridges, together with the funding of Toricor Infrastructure, Windham{-}Waploo, Fort West and Benningon{-}Haterloo.[12]

History [ edit ]

Preconstruction [ edit ]

Main construction [ edit ]

The first casualty of the construction was Lufferon Bridge and Dan Grant Drive. It was closed in 1890. The building was purchased by Buffaloiss, the primary equipment company, in 2003 and built in Centrepount, Ontario.[12]

Work [ edit ]

The original bridge was completed in November 1980. In the 1990s, bridge plans were removed; by 2007, the replacement trench had replaced as the roadway's predecessor.[14] By the 1970s, the bridge was adapted for the water{-}free route, eliminating congestion on thawed trains.[15]

Proposed work [ edit ]

Further reconstruction of the bridge would begin. In 2008, the exposed stones would be removed. In 2011, dredging wire had been installed and a steel beam would be removed.[16] In May 2015, a final masonry scaffolding had been installed and the replacement trench has been removed. The final work was completed by Mississauga in January 2016, about a year after the tunnel was opened.[17]

Mississauga bridge [ edit ]

The Mississauga Bridge is east of the Listed Ontario and First Nation Lines, approximately 55 km (176 mi2) south of the Northwest Territories. It has an approximate walking distance of approximately 700 metres (342 mi2) or beyond the stub crest. Its alignment starts south, follows high/ice descent, and weights less than 50 ft (99 ft) below the invalid edge. The separated Ottawa bridge uses a smooth northpath to pin ice to its high lobe and longitude below its narrow. The bridge faces tendings, which
\end{samplebox}
\clearpage
\refstepcounter{subsection}
\label{sample:replaid-520000-1-nfe128}
\addcontentsline{toc}{subsection}{\protect\numberline{\thesubsection}RePlaID + \ours, 520k: sample 1, 128 NFE}
\begin{samplebox}{\thesubsection\enspace RePlaID + \ours}{Sample 1\enspace\textbar\enspace 128 NFE (+1)\enspace\textbar\enspace GenPPL 27.51}
{\rmfamily\fontsize{7.5}{9}\selectfont\color{sampleink}Unedited model output\enspace\textbar\enspace 520k checkpoint\enspace\textbar\enspace EMA\enspace\textbar\enspace Seed 1\enspace\textbar\enspace Temperature 1\enspace\textbar\enspace 1,024 model tokens\par}\vspace{5pt}
ing while rejecting his country\textquoteright{}s request to sell military hardware to Germany. Firo also vacated the seat of the United Nations Security Council around the time of Truman coming to office\textemdash{}just over 60 years ago. To be fair during his tenure as President, Firo was always functioned in a diplomatic style that was similar to, and policed, by his predecessors.

The last several months of Firo\textquoteright{}s tenure have been a long questioning period for some of his close aides working on military matters and diplomatic affairs, so it\textquoteright{}s hardly surprising that his resignation is taking place in earnest. The most valid reasons for his resignation are that he has a failure to understand current events and that he appears to have a poor grasp for commemorating the 9th century of the Soviet Union. This latter point clearly underlines the importance of Poroshenko\textquoteright{}s resignation and the difficulty some of his close aides face in securing the nomination to be appointed the new United Nations High Commissioner for War, making him only the second Bulgarian to take the throne.

It\textquoteright{}s not clear what effect Poroshenko\textquoteright{}s resignation will have on all of the bilateral relations, but there is no speculation about it. If he is indeed being denied the post, it\textquoteright{}s not difficult to imagine how the world will react with relations with another country in the world before WW2 and after. Something that could lead relations to decline in an destabilizing or very unpredictable manner as a result of Poroshenko\textquoteright{}s resignation. Something else that could set off an uprising against the country with which Poroshenko had become deeply friendly as president, and ultimately lead the special diplomat to live in a very unpredictable manner that speaks against and benefits the country that he was once a friend of.

In this context, it\textquoteright{}s entirely possible that the challenges of the post will soon become insurmountable as at this point a decision decision is in the process of being made.

Republic of Serbia

United States

The Republic of Serbia, as we know it, is an independent political republic established in the Yugoslav Socialist Republic by a former Yugoslav Prime Minister. The Serbian Federal Government has long had a history of dissidence in the form of freedom of movement and freedom of expression. While it is regarded as a peaceful archipelago, it is also the legacy of real development and democracy while at the same time it is beset by poverty and economic challenges. It continues to be a major partner in the cause of justice, peace, and the prosperity and security of human beings around the world.

National Government of the OSCE

The Republic of Macedonia (OSCE) is a legitimate independent political republic established by the former Prime Minister Ivan Milanovi\'{c}. It is now known as the \textquotedblleft{}National Government\textquotedblright{} of the Republic of Macedonia, and it officially includes the Ministry of Trade and Industry of the Republic of Macedonia. From the beginning of its history, it has sought to keep its independence and conserve its natural resources and technology. It is therefore important to describe the origins of this country in simplistic terms.

The Combined Executive Branch is a form of national government from the Croiana Executive Council and Peruvian Regional Government that rules the respective geographic regions of Macedonia and the Adriatic Sea. The Executive Branch oversees the central and national councils in the OSCE at large, which are responsible for setting up the Cabinet of Ministers, Armed Forces, and the army. The Executive Branch also acts as the Office of the Public Prosecutor and ensures that all criminal investigations are conducted in accordance with the applicable Law. The Cabinet also serves Parliament and oversees the rendering of all laws, rights, freedoms, and duties of the people of the country.

Despite its unique flag, the political activities of this country have always been extremely unpopular. It is a little surprising that this country is not known for engaging in illegal activities, but why it\textquoteright{}s quite interesting is that it is a highly controversial political actor.

Republic of Bosnia

Republic of the Yugoslavia is an Eastern Europe{-}based country led by a former Bosnian Member of the Executive Branch of the Yugoslav Socialist Republic. The democratically elected government normally known internationally as the Partisan centenary of the Kosovo Socialist Republic has long had a history of dissidence in the form of freedom of expression.

Prior to the extent of its jurisdiction, a country that is subject to UN sanctions has already been involved in dozens of violent provocations committed for the purposes of complying with the Rule of Association. These incidents have been claimed during popular protests conducted with the support of the armed forces. So while there is no disputing with that assessment why we should believe that this is an independent state operating within the confines of UN humanitarian law, our neighbors might want to take a far closer look at this matter.

The political activities of the Yugoslav National Government have been involved in a series of violent protests carried out to oppose the Serbs\textquoteright{} prolonged rule in Montenegro. These popular demonstrations mobilized ethnic Kosovan workers in the northeastern autonomous regions, and raised concerns regarding their possible ties to the Serbs. These violent protests which were covertly sabotaged and prosecuted with the support of the armed forces have done
\end{samplebox}
\clearpage
\refstepcounter{subsection}
\label{sample:replaid-520000-0-nfe1024}
\addcontentsline{toc}{subsection}{\protect\numberline{\thesubsection}RePlaID + \ours, 520k: sample 0, 1024 NFE}
\begin{samplebox}{\thesubsection\enspace RePlaID + \ours}{Sample 0\enspace\textbar\enspace 1,024 NFE (+1)\enspace\textbar\enspace GenPPL 11.33}
{\rmfamily\fontsize{7.5}{9}\selectfont\color{sampleink}Unedited model output\enspace\textbar\enspace 520k checkpoint\enspace\textbar\enspace EMA\enspace\textbar\enspace Seed 0\enspace\textbar\enspace Temperature 1\enspace\textbar\enspace 1,024 model tokens\par}\vspace{5pt}
 is ready for a big statement to be made. There is no doubt that an exciting time has passed since we challenged our women's to victory in the first round of the championship event. Furthermore, we have another season to go. While representing our country, our eyes will surely be open to the hard work and dedication from these talented individuals. While representing our country, there is no doubt that many of our fans will open their eyes towards our debut at the T{-}Mobile Arena at WrestleMania (sticking to win the title) and watch with the best of our best. We have every reason to be excited for the abilities of these individuals. They have a lot of potential, as they mature to meet our community's expectations. We only have one season to go. Together, we know that this is a very exciting time to come for us here at the T{-}Mobile Arena at WrestleMania. Moreover, there is no doubt that an exciting time has passed since we challenged our women's to victory in the first round of the championship event. Furthermore, we have another season to go and our eyes will surely be open to the hard work and dedication we all continue to see from these talented individuals. With a few matches up next, our upcoming season will be replete with high expectations. Of course, we are going to take some positives out of each win. Members of our roster have continued to impress themselves with consistently strong performances. Through our continued relationship with our partners, we have learned so much while leading these teams. Smackdown only has one season to go. Together, these men and women are beginning to evolve into a wonderful breed. {-} Tres Judge, Turner Broadcasting President

Through our continued friendship and with us as a company, let it be known that we have another season to go for these talented individuals to reach their full potential and return to the table to meet fans expectations. We all know that this is a very exciting time to come for us at UW{-}Meison University and we are very, very excited to see an opportunity at our new organization. While representing our country, an exciting time has passed since we challenged our women's to victory in the first round of the championship event. Furthermore, we have another season to go. While representing our country, there is no doubt that many of our fans will open their eyes towards our debut in the T{-}Mobile Arena at WrestleMania (sticking to win the title) and watch with the best of our best. Furthermore, we have another season to go, for our individuals to reach their full potential and return to the table to meet fans expectations. We all know that this is a very exciting time to come for us here at UW{-}Meison University and we are very, very excited to see an opportunity at our new organization. While representing our country, I would like to state that there is no doubt that many of our fans will open their eyes towards our debut at the T{-}Mobile Arena at WrestleMania. Moreover, we have a lot of work to do, for these individuals to reach their full potential and mature to meet our community's expectations. We only have one season to go. While representing our country, our eyes will surely be open to the hard work and dedication we all continue to see from the air.

While representing our country, we are ready for a big statement to be made at WrestleMania. With two matches up next, an exciting time has passed since we challenged our women's to victory in the first round of the championship event. Furthermore, we have another season to go. Working through our continued friendship and with us as a company, let it be known that our upcoming season will be replete with high expectations. Of course, we are going to take some positives out of each win. Members of our roster have continued to impress themselves with consistently strong performances. Through our continued relationship with our partners, we have learned so much to learn while forming these teams. Smackdown only has one season to go. Together, these men and women are beginning to evolve into a wonderful breed. 31{-}year{-}old Katsuke "Katsuke" Kobakashi is a retired Japanese champion grappler that trains in Brazil, Germany, Italy, France, Poland, Japan, and Australia. He has a wealth of wrestling experience under his belt as he returns once again to the world.

During his successful career, 31{-}year{-}old Kobakashi has lifted over 240 lbs (120 kg) in Japan, Latin America, Australia, the U.K. as well as many other countries. He has competed in the middleweight, super{-}bantamweight, and welterweight middleweight divisions. His promotions include competing in the welterweight, super{-}bantamweight, and bantamweight divisions. He consistently trains around the best elite opponents in the division, including Mauricio Rua, Sasha Banks, and Anthony Pettis. He follows the tradition of Eddie Guerrero, Holly Holm, Wanderlei Silva, Marc Lillachenyk, and Hulk Hogan winning
\end{samplebox}
\clearpage
\refstepcounter{subsection}
\label{sample:replaid-520000-1-nfe1024}
\addcontentsline{toc}{subsection}{\protect\numberline{\thesubsection}RePlaID + \ours, 520k: sample 1, 1024 NFE}
\begin{samplebox}{\thesubsection\enspace RePlaID + \ours}{Sample 1\enspace\textbar\enspace 1,024 NFE (+1)\enspace\textbar\enspace GenPPL 11.33}
{\rmfamily\fontsize{7.5}{9}\selectfont\color{sampleink}Unedited model output\enspace\textbar\enspace 520k checkpoint\enspace\textbar\enspace EMA\enspace\textbar\enspace Seed 1\enspace\textbar\enspace Temperature 1\enspace\textbar\enspace 1,024 model tokens\par}\vspace{5pt}
 as a shortstop, he was a CSIA All{-}Stars and was named the 2014 CSIA East Defensive Player of the Year. Moran led CSIA East to the 2014 Big Star Challenge, where he lost a critical game to Texas. He has appeared in all 61 games of his career at CSIA East and he was inducted into the Hall of Fame.

Moran also began his professional baseball career with the Pizskin department as a utility player and prep player, before academically working as an advisor and designated hitter. Jordan earned a performing arts degree from the University of Nebraska Omaha as a facultyman.

Moving forward, Jordan will serve as an ex{-}center coach with the Royals, who play in the final four seasons of the Negro Leagues American League in Glendale, Ariz. Jordan will also serve as an unrestricted free agent for the next three seasons and is eligible to sign with the Royals, Blue Jays and Diamondbacks.

\textquotedblleft{}I am honored to end my career with the Royals organization,\textquotedblright{} Jordan said in a statement. \textquotedblleft{}I\textquoteright{}m looking forward to ending my great career in this organization and continuing being a part of this organization that is special to me. I am extremely honored to have signed a long{-}term contract with the Kansas City Royals on Monday, June 30, 2017, as I look forward to having a very successful career as a member of this organization. I can now put an end to my successful stints with the University of Nebraska and I look forward to building on this program\textquoteright{}s history and tradition. Professional baseball has always played a key role in developing my game and making my contributions to my sport and this organization. I\textquoteright{}m looking forward to signing a new contract extension and a new long{-}term contract with the Royals and continuing a specific focus on building on my legacy with this team.\textquotedblright{}

Sign up for breaking news, or updates by email. Please stay with us and follow us.

Jordan will serve as an unrestricted free agent for each of the four seasons of the Negro Leagues American Fall League in Glendale, Ariz., until July 15 with the New York Yankees. A non{-}draft option period is currently in effect for the next three seasons. Jordan has served as an ex{-}center coach with Major League Baseball and the University of Nebraska. He will also serve as an unrestricted free agent through the next three seasons and is eligible to sign with the Royals, Blue Jays and Diamondbacks. He also has served as an ex{-}center coach with the Boston Red Sox and the University of Louisville.

Besides serving as vice president of recruiting, director of programming and child development as well as director of scouting, amateur performance and director of prospects, Jordan also served as an assistant coach with the Triple{-}A Palm Beach Suns from 2012 to 2014. Jordan has also served as the senior assistant vice president of operations and communication as well as as a separate assistant vice president of the Triple{-}A Gulf Coast League.

Moran has spent large portions of his career with the Pizskin department playing in the Coca{-}Cola League and a large portion of his career in the Crosman League.

Advertisement

Redbirds co{-}owner and Red Sox broadcaster John Farrell:

\textquotedblleft{}I am very excited to have a new opportunity to join this organization and I want to continue my development as a person who has built incredible relationships with Pizskin players and coaches throughout the organization at all levels. I am looking forward to bringing competitive baseball to Kansas City and also an enormous honor to work with the Royals, the Twins and the Boston Red Sox. Thank you for your commitment to the Kansas City organization and this organization has been a great home to my career.\textquotedblright{}

The 25{-}year{-}old Moran, who returned from camp in January to serve as a bullpen coach with the Royals, will play in the Negro Leagues American League in Glendale, Ariz., until July 6, and will fly back to Washington, D.C., the following day. Jordan will also serve as an unrestricted free agent for each of the next three professional seasons with the New York Yankees; a non{-}draft option period is currently in effect for the next three seasons.

\textquotedblleft{}I am honored to end my career with the Royals and be a part of this organization that is special to me,\textquotedblright{} said Royals co{-}owner and Red Sox president Alex Shapiro. \textquotedblleft{}Moving forward I\textquoteright{}m confident that I will be looking forward to building on my Hall of Fame career.\textquotedblright{}

Sign up for breaking news, or updates by email. Please stay with us and follow us.

Jordan will serve as an ex{-}center coach with the Royals, who play in the final four seasons of the Negro Leagues American Fall League in Glendale. Jordan will also serve as an ex{-}center event leader for the Royals for Tuesday night\textquoteright{}s regular season finale. Pizskin senior vice president Matt Hantiss will manage the Royals\textquoteright{} team.

Moran will remain with the Royals through his first season with the Blue Jays and will join the class of a future Louisville Slugger Hall of Fame member.

Advertisement
\end{samplebox}

\end{document}